\documentclass[10pt,journal,compsoc]{IEEEtran}

\usepackage{graphicx}

\usepackage{wrapfig}

\usepackage{color}
\usepackage{float}
\usepackage{hyperref}
\definecolor{mscolor}{rgb}{0,0,0.8}

\definecolor{cmcolor}{rgb}{1,0,0.8}

\definecolor{bfrcolor}{rgb}{1,0,0}

\definecolor{abcolor}{rgb}{0,0.5,0.4}

\usepackage[labelfont=sf, textfont=sf, compatibility=false]{caption}
\usepackage[labelfont=sf, textfont=sf, justification=raggedright]{subcaption}

\usepackage{amsmath}

\usepackage{microtype}                 
\PassOptionsToPackage{warn}{textcomp}  
\usepackage{textcomp}                  
\usepackage{mathptmx}                  
\usepackage{times}                     
\usepackage{cite}                      
\usepackage{tabu}                      
\usepackage{booktabs}                  

\begin{document}

\title{DumbleDR: Predicting User Preferences of Dimensionality Reduction Projection Quality}


\author{Cristina~Morariu,
        Adrien~Bibal,
        Rene~Cutura,
        Beno\^it Fr\'enay,~\IEEEmembership{Member,~IEEE} and
        Michael~Sedlmair,~\IEEEmembership{Member,~IEEE}
\thanks{Cristina~Morariu and Michael~Sedlmair are with the University of Stuttgart.\protect\\ Email: \{cristina.morariu, michael.sedlmair\}@visus.uni-stuttgart.de.}
\thanks{
Rene~Cutura is with the University of Stuttgart and TU Wien. \protect\\Email: rene.cutura@tuwien.ac.at.}
\thanks{
Adrien~Bibal and Beno\^it Fr\'enay are with the University of Namur. \protect\\Email: \{adrien.bibal, benoit.frenay\}@unamur.be.}
\thanks{
Cristina Morariu and Adrien Bibal are joint first authors.}}



\IEEEtitleabstractindextext{%
\begin{abstract}
A plethora of dimensionality reduction techniques have emerged over the past decades, leaving researchers and analysts with a wide variety of choices for reducing their data, all the more so given some techniques come with additional parametrization (e.g. t-SNE, UMAP, etc.). Recent studies are showing that people often use dimensionality reduction as a black-box regardless of the specific properties the method itself preserves. Hence, evaluating and comparing 2D projections is usually qualitatively decided, by setting projections side-by-side and letting human judgment decide which projection is the best. In this work, we propose a quantitative way of evaluating projections, that nonetheless places human perception at the center. We run a comparative study, where we ask people to select ‘good’ and ‘misleading’ views between scatterplots of low-level projections of image datasets, simulating the way people usually select projections. We use the study data as labels for a set of quality metrics whose purpose is to discover and quantify what exactly people are looking for when deciding between projections. With this proxy for human judgments, we use it to rank projections on new datasets, explain why they are relevant, and quantify the degree of subjectivity in projections selected. 
\end{abstract} 

}

\IEEEdisplaynontitleabstractindextext

\maketitle


\IEEEpeerreviewmaketitle

\section{Introduction}\label{sec:intro}

\IEEEPARstart{A} wide-spread approach for data exploration is the use of dimensionality reduction (DR) techniques, also known as projections. DR is a process that projects high-dimensional data to a lower-dimensional space, such that the resulting projection retains specific properties from the original data. An application of DR is in visualization, where users can create scatterplots based on two retained dimensions as part of their data analysis. DR methods are used in various domains ranging from biology and medical research to social sciences, and they are actively researched in both the machine learning (ML) and visualization (VIS) communities. 

An extensive amount of techniques exists to produce such projections, such as principal component analysis (PCA)~\cite{Bishop2006}, multidimensional scaling (MDS)~\cite{kruskal1964NMStress}, isometric feature mapping (Isomap or ISM)~\cite{tenenbaum2000ISOMAP}, $t$-distributed stochastic neighborhood embedding ($t$-SNE)~\cite{vdm2008tSNE} and, more recently, uniform manifold approximation (UMAP)~\cite{mcinnes2018umap}. These methods can produce widely different results, all the more so given that some have hyper-parameters (e.g. the perplexity of $t$-SNE). 

Evaluating the quality of these results is, however, the burden of users. In a typical process, a user generates a range of projections, visualizes them in scatterplots, and selects a suitable one from the line-up.
Several attempts have been made to improve our understanding of what users look for when evaluating projections. Some studies focus on investigating whether human judgment is indeed reliable for evaluating projections \cite{lewis2012behavioral}, while others focus on defining the tasks that users perform when investigating projections \cite{brehmer2014visualizing}. Previous works also show that people use DR as a black-box mechanism without necessarily understanding what the objective of the specific technique is \cite{lewis2012behavioral,lewis2012ABW}. To consolidate the evaluation of projections quantitatively, both the ML and VIS communities proposed quality metrics that can be used to select the best projections automatically. 

In this paper, we aim at bridging previous research on quality metrics for dimensionality reduction and scatterplot visualization, with the work done on understanding human judgments of projection quality. We evaluate to what extent existing metrics in the literature can quantify user preferences.
To this end, we gathered collections of images that we used to compute widely-used DR techniques. 
In total, 11 image collections were used, and 25 projections were computed, resulting from different parametrizations of the DR techniques mentioned above. Based on this data, we ran a 54 person user study to collect preferences on these projections. We then investigated in how far these human preferences can be formally expressed through existing quality metrics.
Our aim is thus not to survey all DR methods, but rather to investigate whether quality metrics, or a combination thereof, can capture user preferences.

Our problem can be framed as a supervised learning problem, where the relationship between a combination of various quality metrics is used to predict human judgments. To solve this problem, machine learning models are used to compute how these metrics should be combined. The aim is to create and provide a model that can both predict projections users would most likely prefer, as well as to offer an explanation as to why they prefer them.

There are two main reasons for this choice.
First, building a supervised model will allow us to derive a composite metric based on user perception. The new metric can then be used to select projections that would generally be considered interesting. This is specifically important when many DR techniques are considered, or for DR techniques that have several non-trivial hyper-parameters to tune. Second, this approach will  enable us to compare which quality metrics are important for expressing human preferences. 

In summary, our work makes the following contributions: 
\begin{itemize}
\item the collection and analysis of data from a 54-participant user study on subjective preferences in DR projections;
\item a quantitative analysis that (a) explains what users like when selecting DR projections, (b) sheds light on the  feasibility of predicting preferences with quality metrics, and (c) allows us to better understand which ML and VIS metrics are important for that;
to that end, we use three modeling approaches that allow to combine quality metrics to predict user preferences of projections on unseen data, as well as an analysis on which approach performs best;
\item DumbleDR, a proof-of-concept web tool that uses the best-performing model to rank projections for new datasets and show what metrics drove the ranking of the results. 
\end{itemize}

\section{Background \& Related Work}\label{sec:SotA}

Our work brings together the two main types of evaluation in dimensionality reduction (DR): the quantitative evaluation using visual and DR-specific quality metrics, and the qualitative evaluation based on human judgments. This section presents the latest work in these two areas, and explains how our contributions build on top of this knowledge. 

\subsection{DR Evaluation using Quality Metrics}\label{subsec:RW_QM}

Measuring the quality of projections is the work of two communities, and each brought quality measures that have distinct properties. These different quality metrics are presented in this section. 

\subsubsection{Measures from the Machine Learning Community}

The machine learning (ML) community has defined several measures that can be used as objective functions within DR algorithms. A good example is \emph{stress}, the well-known objective function of multidimensional scaling, which measures the preservation of pairwise distances between the instances in the high-dimensional (HD) and the low-dimensional (LD) spaces. 
Beyond that, the ML community has investigated metrics that seek to define and measure the quality of the DR process itself. 
The rationale for this choice is that metrics that are used in objective functions are constrained in their definition (e.g. being differentiable), constraints that may not be necessary if the sole purpose is to measure quality~\cite{lee2010}. 

Examples of such measures are the local continuity meta-criterion (LCMC)~\cite{chen2009}, the measure of trustfulness and continuity (Truthfulness and Continuity)~\cite{venna2006} and AUC$_{log}$RNX~\cite{lee2015}. These measures typically check if the neighborhoods in the HD space are preserved in the projection.
For instance, LCMC computes, for each point, the average number of neighbors it has in common in HD and LD for a certain neighborhood size $k$. 
\emph{Truthfulness}, on the other hand, is defined by roughly summing the rank of all pairwise distances from a point $i$ in the original HD data to its nearest neighbors in the LD projection that are not among the $k$ nearest neighbors of $i$ in the original data. This metric seeks to measure whether one can trust what can be seen in the visualization. The measure of \emph{continuity} is the exact opposite, as it tells how well the patterns from the original dataset are projected in the visualization. The Continuity for a particular neighborhood size $k$ is defined by the rank of all pairwise distances from the point $i$ in the LD projection to the nearest neighbors of $i$ in the original HD data that are not among the $k$ nearest neighbors of $i$ in the LD projection.
While the previously mentioned approaches focus on a specific neighborhood size $k$, AUC$_{log}$RNX consider all neighborhood sizes, with a focus on smaller neighborhoods. In order to do so, AUC$_{log}$RNX considers, for each point, the number of neighbors in common in LD and HD for all neighborhood sizes with a logarithmic importance.

\subsubsection{Measures from the Visualization Community}

The other community that tackles measuring projection quality is the visualization (VIS) community. Metrics from the VIS community generally focus on the quantification of visual patterns projections/scatterplots. A venerable example of such measures are the Scagnostics measures~\cite{wilkinson2005,wilkinson2006}), that quantify patterns such as Sparsity,  Skewness, and Outlierness.

Recently, a substantial amount work has focused on measuring class separability, that is, how well classes are separated in a DR projection.
Distance consistency (DSC), for instance, computes the number of instances that are closest to the centroid of their own class rather than another class. Alternatively, SepMe~\cite{aupetit2016sepme} provides an ensemble of separability metrics that use neighbourhood graphs to assess how well classes are separated. These metrics are currently the best performing separability metrics evaluated in literature. 

Other popular measures in this category are the average between-within clusters (ABW)~\cite{lewis2012ABW}, the hypothesis margin (HM)~\cite{gilad2004HM}, the neighborhood hit (NH)~\cite{paulovich2008NH} and the Calinski-Harabasz index (CAL)~\cite{calinski1974CAL}. All these metrics measure the separability between clusters, albeit differently. 

Similar to our goals, several recent works \cite{pandey2016towards, bertini2011quality, lehmann2015study, etemadpour2014perception, aupetit2016sepme} focused their attention on evaluating quality metrics against human perception, although with different use cases. Sedlmair and Aupetit~\cite{aupetit2016sepme, sedlmair2015quality} examine perception of class separability in color-coded scatterplots, Pandey et al.~\cite{pandey2016towards} assess to what extent Scagnostics can be used as a proxi for human perception, and Lehmann et al.~\cite{lehmann2015study} evaluate whether Scagnostics can be used to filter perceptually interesting views for users. 
None of these works, however, has focused on recommending DR methods and explaining this recommendation using quality metrics, as we do.

\subsubsection{Accuracy and Interpretability Measures}

The main difference between the measures designed in ML and those in VIS is their objective. ML metrics generally seek to measure how well the information is preserved when reducing the dimensions. In contrast, VIS metrics  tend to focus on the presence of patterns in the visualizations that make it possible for users to grasp their visualizations and get insights about their data. Following the parallel of Bibal and Fr{\'e}nay~\cite{bibal2019ICLR} with supervised learning, the ML measures would be ``accuracy'' measures, while VIS measures would be ``interpretability'' measures. And, as in supervised learning, the two types of measures should be balanced to obtain results that would satisfy users~\cite{bibal2016NIPS,bibal2019ICLR}. Indeed, accuracy measures are necessary because visualizations with well-separated clusters are not useful if they are not faithful to the high-dimensional space. Likewise, interpretability measures are also necessary as if readable patterns are not provided, nothing may be taken from the visualization.

\subsubsection{Combining the Different Quality Measures}

One idea, which is the one followed by this paper, is to combine the two worlds by mathematically combining the metrics. For instance, Bibal and Fr{\'e}nay~\cite{bibal2019ICLR} formulated the linear combination of quality metrics as follows:
\begin{equation*}\label{alpha_beta}
\begin{aligned}
\text{combination} = & \text{ } (\alpha_1 * AM_1) + ... + (\alpha_i * AM_i) + ... + (\alpha_m * AM_m) \\ 
          & + (\beta_1 * IM_1) + ... + (\beta_j * IM_j) + ... + (\beta_u * IM_u),
\end{aligned}
\end{equation*}
where $AM$ (resp. $IM$) means accuracy metric (resp. interpretability metric). The different $\alpha$ and $\beta$, which are learned, represent the contribution of the metric to which they correspond.

Ensembles of metrics were also discussed in the quantitative survey of DR methods of Espadoto et al.~\cite{espadoto2019towards}. The authors surveyed 44 DR methods and computed the average of several metrics (truthfulness, continuity, neighborhood hit, normalized stress, Shepard goodness and local error) on 18 datasets in order to assess the global performance of individual DR techniques. We build on this work and go beyond by investigating learning the combination of measures that predict user choices. Similarly, Nonato and Aupetit~\cite{nonato2018multidimensional}, as well as van der Maaten et al.~\cite{van2009dimensionality}, extensively reviewed DR techniques alongside quality metrics for DR, albeit without computing quality metrics on projections.

\subsubsection{Applications for Quality Metrics}
Aside from the works mentioned above, the VIS community focuses on bridging the gap between quality metrics and human judgments by designing visual analytics (VA) systems that aid users in comparing~\cite{cutura2020comparing} or selecting~\cite{cutura2018viscoder, ingram2010dimstiller, martins2014visual} projections. The insights derived from our contribution can be used as part of a VA system that recommends projections. 

Lehman et al.~\cite{lehmann2015study} also propose using specific quality metrics to automatically filter out easily rejected projections, as scored by users. Wang et al.~\cite{wang2017perception} use previously evaluated quality metrics of subjective class separability to propose a new DR technique, which is implicitly optimized to model human perception of separability. 

\subsection{Evaluation Driven by Human Judgments} 

Despite the existence of quality metrics, the burden in the evaluation of projections remains mainly on users. This section discusses DR research that collects and/or uses human judgment to assess quality.

\subsubsection{Taxonomies for high-level tasks related to DR}
The work by Brehmer et al.~\cite{brehmer2014visualizing} aims to define what tasks users perform when they investigate projections. Following interviews, the authors introduce a characterization of tasks. These are \textit{manifold tasks}, where users are trying to name the synthesized dimensions, and \textit{cluster tasks}, where users verify, name, or match clusters with class names. These tasks have been considered in the selection of our datasets to ensure our study participants deal with different settings. Another closely aligned work is the one of Sedlmair et al.~\cite{sedlmair2012taxonomy}, which proposes a cluster analysis taxonomy, one of the most important analysis tasks in the DR data exploration process. 

\subsubsection{Assessing user preferences in DR}
Lewis and van der Maaten~\cite{lewis2012behavioral} investigate whether human judgments are consistent by running a user study with groups of experts and novices. The participants are asked to select 2 good projections and a bad one from a line-up of 9 monochrome scatterplots, each representing a projection. They offer the users little information regarding the original dataset and find out different users prefer different projections, inferring that user preferences are vastly subjective. However, they also show that the more users have expertise, the more they are coherent in their judgement. Our study setup builds up on this one, as both studies focus on the real-life task of users selecting projections from a line-up. However, our goal is (i)
~to deepen the understanding about how users make their decisions and (ii)~to model these for recommending projections. Our setup is detailed in Section~\ref{sec:data_collection}.

Bibal and Fr{\'e}nay~\cite{bibal2016NIPS} also ran a user study collecting user preferences of $t$-SNE projections of the MNIST dataset. The objective of the authors was to study how cluster separability measures and their combination (using a modified Cox model) could predict user preferences. The study presented in this paper is larger in scale at all levels: more datasets, more DR techniques (not only $t$-SNE), more quality metrics and different ways to frame the problem and to combine metrics. This enlargement in scope allows us to perform original analyses and to draw insightful conclusions.

\subsubsection{Selecting DR projections}
Oftentimes, when new DR methods are introduced, a comparative study to other techniques is proposed as an evaluation. The projections get visualized in scatterplots and the reader is invited to assess the line-up and decide for themselves which is the superior projection. This can also be the case for the selection of hyper-parameter values inside a particular DR technique. For instance, the authors of $t$-SNE invite users to try various parametrizations and select the projection they prefer
~\cite{vdm2008tSNE}. 

Wattenberg et al.~\cite{wattenberg2016how} show that blindly trying hyperparameters and selecting appealing projections has downfalls, in that it can mislead users on the faithfulness of the projection. Moreover, user guidelines given by authors often are technique-specific, in this particular case, for $t$-SNE. To overcome such issues, Sedlmair et al.~\cite{sedlmair2013empirical} assess the best visualization methods to use during DR exploration, and provide guidelines on selecting DR techniques using visualizations based on data collected in a user study. 

Other work~\cite{etemadpour2014perception} designed a user study to assess which projections can best enhance users' abilities to detect clusters, outliers or estimate density. These results were, however, not used to recommend better projections for specific tasks. 

\begin{table*}[!ht]
    \centering
    \scriptsize
\begin{tabular}{l|l|l|c}
     \textbf{Dataset Name} & \textbf{Description}  & \textbf{Difficulty (as scored by users)}& \% of Disagreement\\
     \hline
     COIL-100         & Images of common objects photographed from different angles (128 x 128)    & \includegraphics[height=0.4cm]{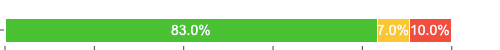}& 11\%  \\
     MNIST       & Handwritten Digist (28 x 28)  &  \includegraphics[height=0.4cm]{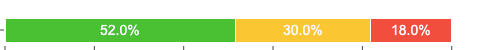}& 12.5\% \\
     Fashion MNIST   & Images of clothes (28 x 28) & \includegraphics[height=0.4cm]{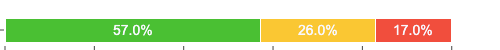}& 20\% \\
     
     Stanford Faces         & One bust photographed from different angles, in different light conditions (50 x 50) &  \includegraphics[height=0.4cm]{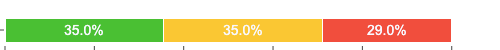}& 19\% \\
     Yale Faces         &  14 people displaying happy, neutral or sad faces  (320 x 243)  &  \includegraphics[height=0.4cm]{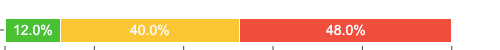}& 20\% \\
     Flowers         &  Photos of 6 different species of flowers    (500 x 500) & \includegraphics[height=0.4cm]{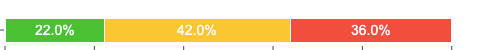}& 20\% \\
     Caltech plants &  Photos/illustrations of 6 different species of plants   (320 x 243) &  \includegraphics[height=0.4cm]{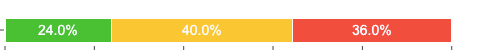}& 18\%  \\
    Caltech vehicles &  Photos/illustrations of 6 different types of vehicles  (320 x 243) &  \includegraphics[height=0.4cm]{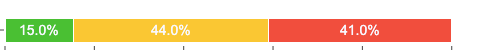}& 22\% \\
    Caltech instruments &  Photos/illustrations of 6 different types of instruments (320 x 243) & \includegraphics[height=0.4cm]{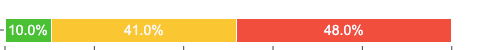}& 21\%  \\
    Paris Buildings & Photos of buildings in Paris (1024 x 768) & \includegraphics[height=0.4cm]{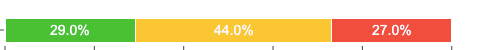}& 14\%  \\
     Oxford Buildings       & Photos of attractions in Oxford  (1024 x 768) & \includegraphics[height=0.4cm]{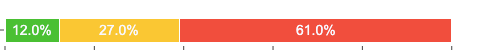}& 24\%  \\
\end{tabular}

\caption{This table lists the datasets used in our experiment. The name, description, the proportion of difficulty ratings given by users (easy - green, medium - amber, hard - red) and the amount of preference disagreements for each dataset are provided as scored by the users. 
}
\label{table:datasets}
\end{table*}

\section{User Study \& Data Collection}\label{sec:data_collection} 

The main idea behind our approach is to (a) sample different DR projections from a set of datasets, (b) collect human DR preferences for them, and (c) calculate different quality metrics to see how far they can predict human DR preferences.
In the following, we describe these three components in more details:

\subsection{Selected Datasets \& Projection Methods}\label{subsec:images_projections}

In the first step, we need to select suitable datasets that allow users to provide meaningful quality judgments for different DR projections thereof. To be able to extract meaningful preferences of projections from users that go beyond the appealing aspect of scatterplots, one needs to make sure that users can process the high-dimensional data they are analyzing as well. 
From the work of Lewis et al.~\cite{lewis2012behavioral}, we know that assessing preferences by only supplying minimal information about the original data can result in highly subjective and inconsistent judgments across participants. It might not be possible to properly judge whether meaningful clusters appear or whether a manifold was adequately unrolled~\cite{brehmer2014visualizing}. 

\begin{figure*}[ht]
    \centering
    \begin{subfigure}[c]{.45\linewidth}
        \includegraphics[width=\linewidth]{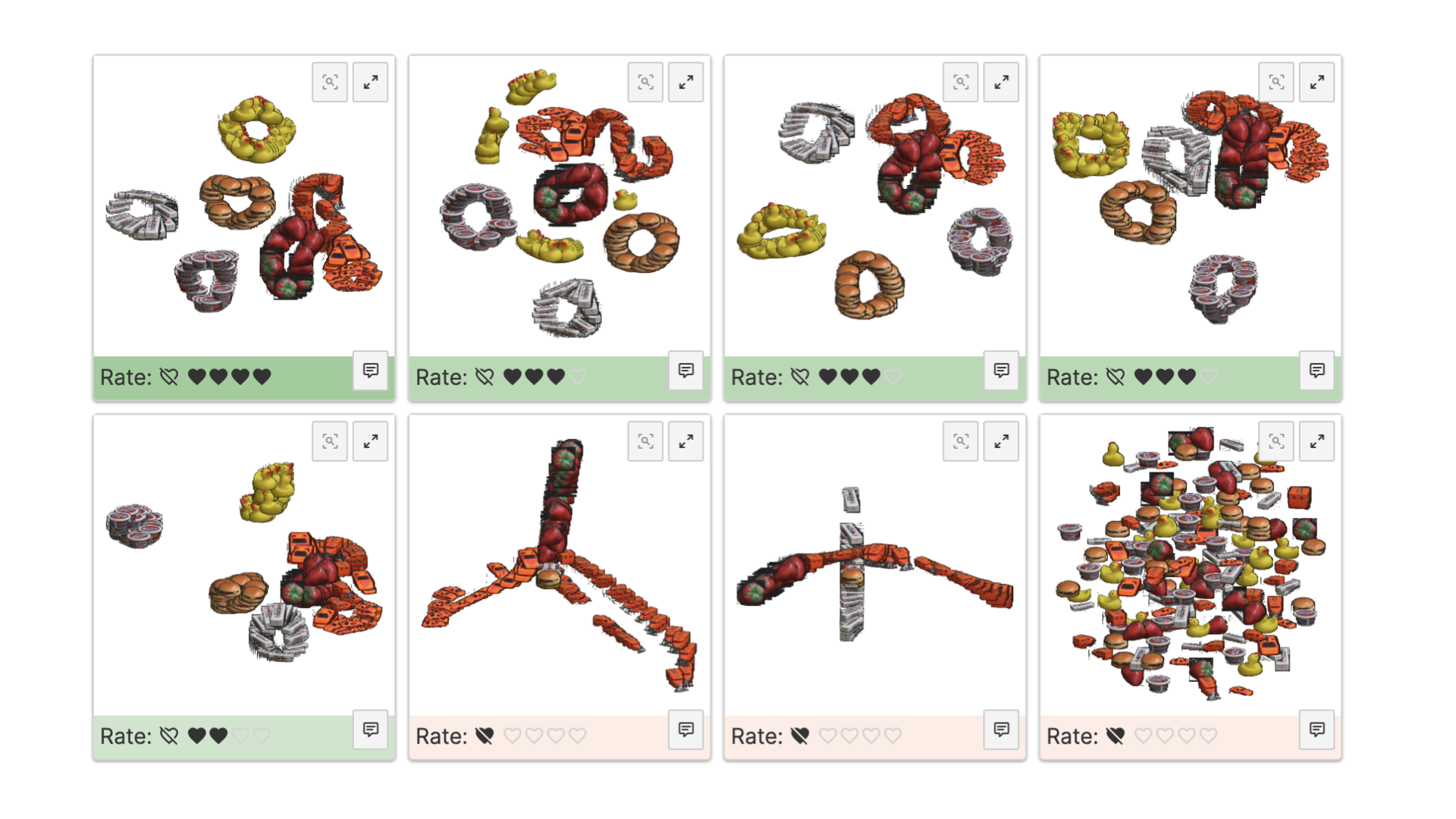}
        \subcaption{Image scatterplot view of the interface. This view is used so that users can see, through thumbnails, how the images from the dataset have been projected in 2D.}
        \label{fig:triala}
    \end{subfigure}
    ~
    \begin{subfigure}[c]{.45\linewidth}
        \includegraphics[width=\linewidth]{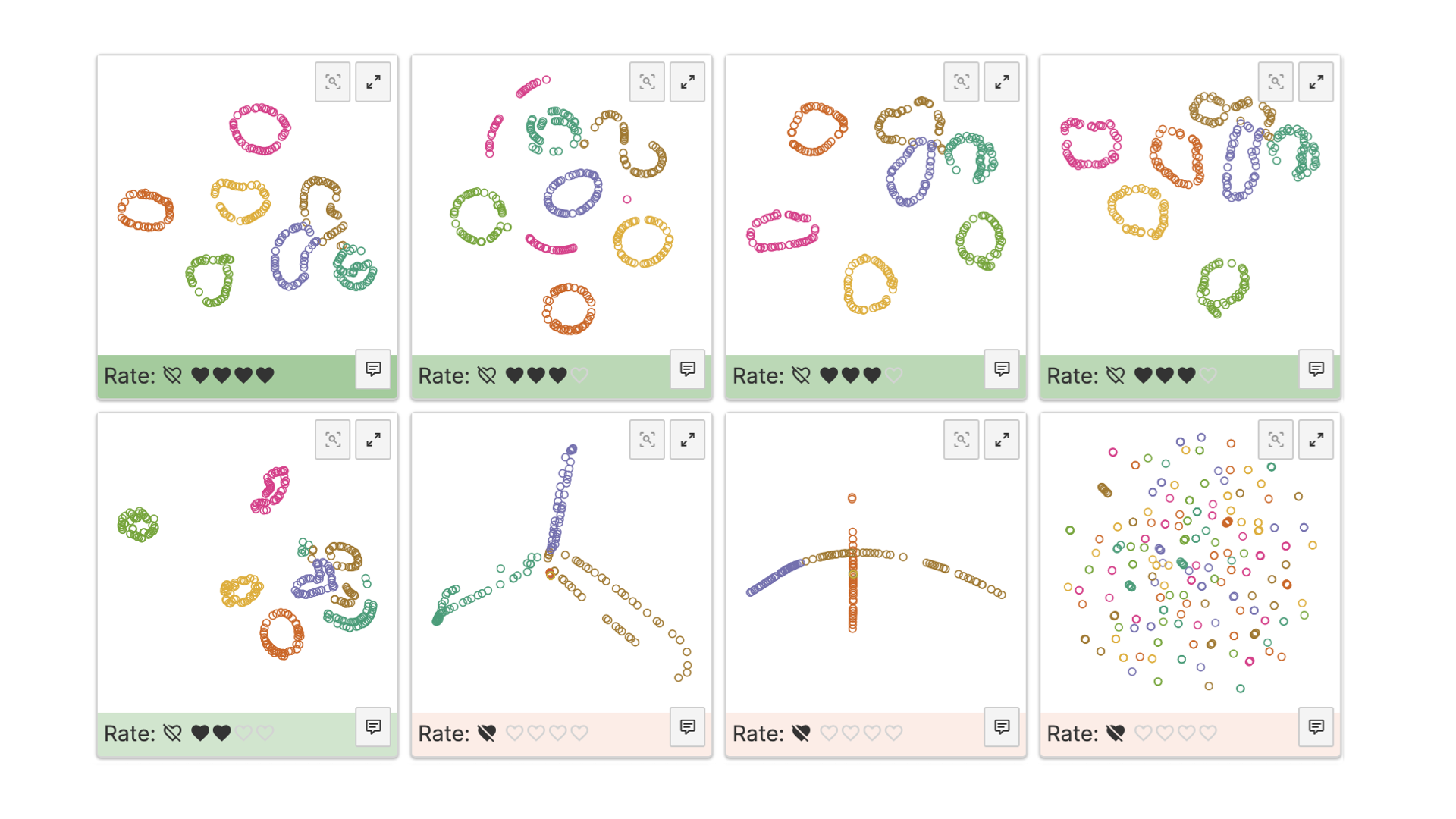}
        \subcaption{Point scatterplot view of the interface. This simpler view contains points instead of image thumbnails, with colors corresponding to class labels.}
        \label{fig:trialb}
    \end{subfigure}
    \caption{Two views of the same trial from the experiment for collecting user preferences. Each view contains 8 projections of COIL-100 built by different DR techniques. Black hearts correspond to the scores distributed among the projections.}
    \label{fig:trial}
\end{figure*}

To solve this issue, we decided to use collections of images for our study. Under this setup, the projections visualized as scatterplots would not be simply monochrome scatterplots. Instead, each dot encoding a 2D position was replaced by a thumbnail of the image getting projected at this location. For example, in the case of the COIL-100 dataset, a collection of objects photographed from different angles, the scatterplot contained thumbnails of objects as shown in Figure~\ref{fig:triala}. By showing images as thumbnails, an access to the high-dimensional attributes (the pixels) is given along with the projected low-dimensional position in the visualization.

We collected a total of 11 image datasets, listed in Table~\ref{table:datasets}. 
First, we selected datasets that implicitly suggest different potential tasks even though no task is explicitly defined in the experiment. For example, in the case of the MNIST digits dataset, the expected task was matching class names (the digits) to various clusters formed. In contrast, for the Stanford face dataset where a bust is photographed from different angles and at different lighting conditions, users could prefer a manifold where the lighting goes from light to dark, or one where the view angle changes smoothly. Second, we sought to collect datasets of various difficulties, on the premise that it is much easier to state a preference on projections from an easy dataset like MNIST, as opposed to a more complex dataset like the Paris Building dataset consisting of larger and more messy real-world photos. We used the original image size (in the column ``Description'' of Table~\ref{table:datasets}) as a proxy measure of dataset complexity, and during the study, users were asked to score the dataset difficulty. For each dataset, its difficulty, aggregated from user responses during the study, is conveyed in Table~\ref{table:datasets}.

The dimensionality reduction techniques used to generate the projections are principal component analysis (PCA)~\cite{Bishop2006}, multidimensional scaling (MDS)~\cite{kruskal1964NMStress}, isometric feature mapping (Isomap)~\cite{tenenbaum2000ISOMAP}, $t$-distributed stochastic neighborhood projection ($t$-SNE)~\cite{vdm2008tSNE}, uniform manifold approximation (UMAP)~\cite{mcinnes2018umap}, locally linear projection (LLE)~\cite{roweis2000LLE}, Spectral Embedding (SE) \cite{ng2001spectral}, and Gaussian random projection (GRP)~\cite{bingham2001GRP}. For techniques with hyper-parameters, multiple projections were generated. One hundred projections were initially generated for each dataset and, then, 25 projections for each dataset were uniformly sampled based on the metric space to be used in the user experiment. Finally, we manually down-sampled projections that appeared very similar, e.g. rotated variants, or duplicates of one another. This process resulted in 15 to 20 distinct projections per dataset. An example of the selected projections can be seen in Figure~\ref{fig:trial}. 
The  parameter settings of Isomap, LLE, $t$-SNE and UMAP can be found in Figure~\ref{fig:dr_descriptive} (right side).
In the next step, we showed these projections to users in an online experiment as detailed in the following section. 

\subsection{User Preferences Dataset}\label{subsec:coll_preferences}

This section describes the user experiment that has been set up to collect user preferences on projections.

\subsubsection{Participants}
In total, 54 users participated in our study, out of which 4 had finished a Ph.D., 38 had a master's degree and the remainder 12 had completed a bachelor's degree. We reached our user base by advertising the study within the university network of the co-authors. Participation was voluntary and unpaid. We asked participants for their domain expertise in machine learning, visualization and dimensionality reduction, and the majority of our user base reported familiarity with all these concepts. 

\subsubsection{Study Procedure}
We conducted a web-based user study that takes place completely online and on various display sizes. The study began with an information page explaining the subject of the study, and the duration it takes (40 to 60 minutes). The participant could only access the study if their display size was larger than 700 x 500. After reading the information page, the users were presented a consent form, a general introduction explaining what dimensionality reduction is and how the user interface of the study works, and with a questionnaire to collect demographic and experience data (see above). The study then proceeded with the trials. Upon finishing, participants were asked about the overall difficulty of the setup and any other feedback.

\subsubsection{Trial Setup}
Our study consisted of multiple trials in which users had to rate projections. The stimuli in each trial were the projections generated by applying DR algorithms to the aforementioned datasets. 

A total of 8 projections of the same dataset were shown per trial. The projections were randomly selected from the total projections available for a particular dataset and were placed on a 2-by-4 grid in a randomized order. The DR projections were shown as scatterplots of images on a white background. The views were connected by brushing and linking, so if a user hovered over one image within a scatterplot, this became highlighted across all eight plots. Users could also enlarge and zoom in one particular view. 

At the beginning of each trial, participants received 15 points (represented by hearts in the interface) and were asked to distribute them across the eight projections. A higher number of points assigned to a projection means that the participant preferred this projection more. One projection could receive a maximum of 4 points. A user could also mark a projection as bad, rather than distributing any points to it. Participants could also sort the grid of DR projections such that  they got rearranged by preference in descending order.  Sorting enabled the participants to focus on a local comparison of projections with their slightly better and worse direct neighbors. The sorting mechanism together with the restricted number of points per trials were designed to force the users into deciding which projections they liked more and which not. Our intention was to avoid a situation where a user would award every projection an equal number of points.  A rated and sorted example of a trial is presented in Figure~\ref{fig:trial}. 

Upon completion of one trial, participants were asked to score the difficulty of the trial and whether they would like to score another dataset. Each user could complete up to 10 trials, each trial testing a dataset. The datasets across trials appeared in random order. Sampling with replacement was used to choose the next trial, meaning a user was able to see the same dataset twice, but with a different selection of projections. The setup was implemented using a serverless architecture in JavaScript\footnote{The user study is available here: \url{https://kix2mix2.github.io/DumbleDR/public/index.html}}. The data collected during the setup was hosted in Germany. 

\subsubsection{Descriptive Results}

An important aspect to analyze was the degree of consensus between users when it came to preferences. Previous work~\cite{lewis2012behavioral} showed that there was a high degree of subjectivity when it comes to users recording preferences of DR projections. Furthermore, users' ability to select good quality projections was called into question. In our study, however, we report that while there were disagreements in ratings, the majority converged towards well-defined preferences. For each pairwise comparison between two projections, the best case scenario is that all judgments are in agreement, i.e. 0\% of disagreement. The worst case is that opinions are evenly split when comparing the projections, i.e. half of the judgments are in disagreement with the other half (50\% of disagreement). In our case, on average, 18.5\% of the ratings were in disagreement with the majority. 
A breakdown of disagreement in conjunction with the difficulty of the dataset as scored by the user can be seen in Table~\ref{table:datasets}. Datasets perceived as harder also incurred a higher percentage of disagreements. One example is the dataset of building photos from Oxford. The same applies the other way around, where ``easy'' datasets such as MNIST had low percentage of disagreements.   

\begin{figure}[ht]
    \centering
    \includegraphics[width=\linewidth]{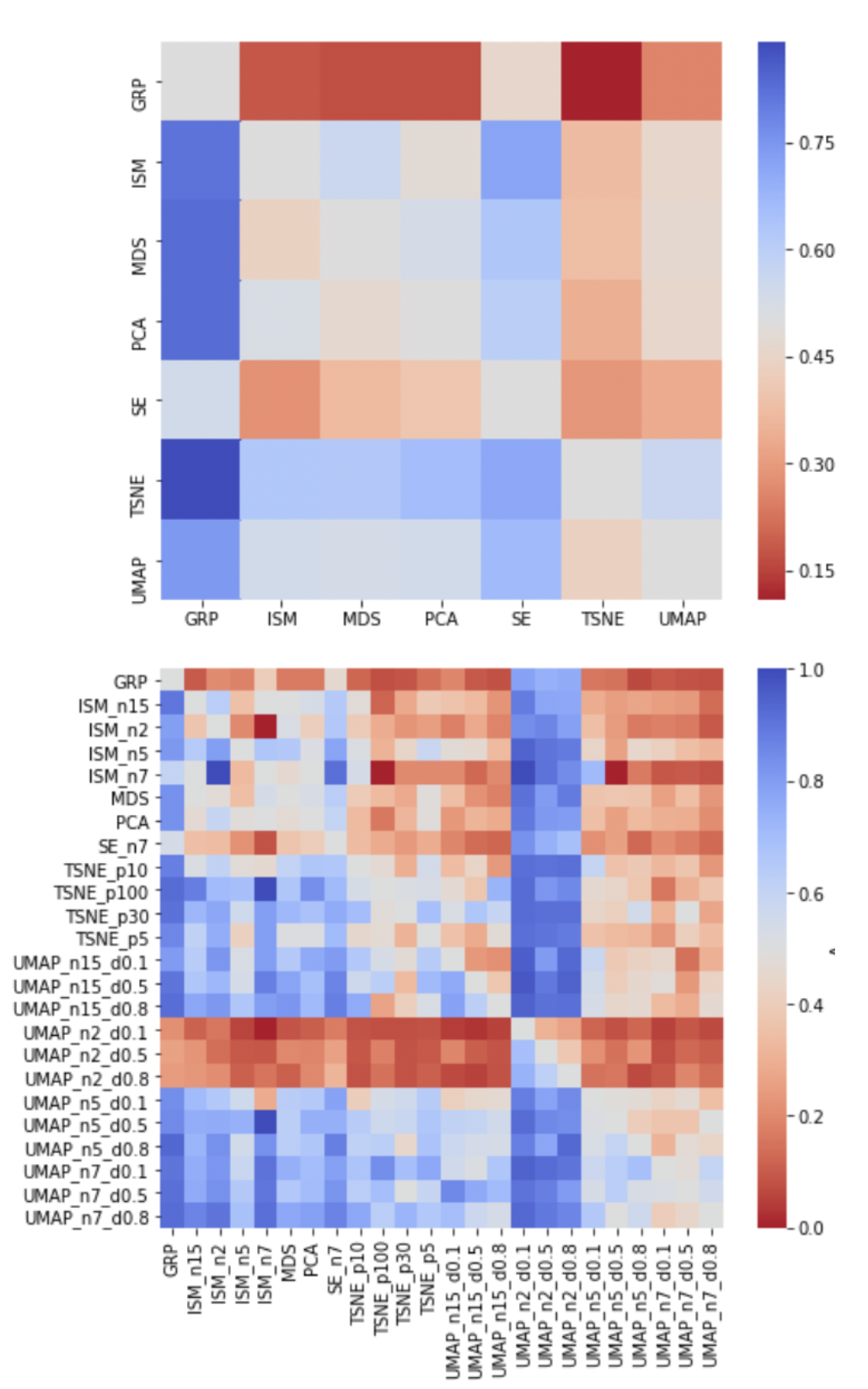}
    
    \caption{User aggregated preferences of DR technique, overall (first) and parametrized (second). A score higher than 0.5, depicted in blue, means that more than 50\% of the users preferred the DR technique specified in the row over the one specified in the column. Scores lower than 50\% are encoded in red. The DR techniques are sorted by user preference in ascending order, with the exception of some parametrizations of UMAP which are universally disliked.
    }
    \label{fig:dr_descriptive}
\end{figure}

Based on the ratings awarded in each trial by each user, we calculated a preference matrix by counting how many times a projection was scored higher than another one. We aggregated these results to assess whether particular DR techniques were systematically preferred. In Figure~\ref{fig:dr_descriptive}, we can see the user preferences aggregated on a DR technique level. The heatmap encodes how many times users agreed that one DR technique (mentioned row-wise) was better than another (column-wise). The bluer the cell the more people agree that the DR technique in the row was better than a technique in the column. There are clear winners and losers. For example, the Gaussian Random Projections (GRP) were universally disliked alongside bad parametrizations of UMAP (e.g. when only two neighbours are considered). This may indicate that GRP could be used as a baseline in further experiments. Interestingly, there were no universally bad parametrization for $t$-SNE. UMAP with good parametrizations appeared to be systematically preferred over the other projections. A hierarchy can be observed: GRP $\leq$ PCA $\leq$ Isomap $\leq$ $t$-SNE $\leq$ UMAP, where $DR_i \leq DR_j$ means that the visualizations generated by $DR_j$ were more often preferred to the visualizations generated by $DR_i$.

\subsection{Quality Metrics Dataset}\label{subsec:coll_metrics}

In order to predict user preferences, we gathered metrics from different communities that measure various aspects of visualizations. The measures quantify both the accuracy and the interpretability of visualizations as defined in Bibal and Frénay~\cite{bibal2019ICLR}. All metrics were normalized such that their value is between 0 and 1. The list of metrics used, as well as whether they measure the correctness of the HD-to-LD mapping, or the quality of the LD visualization only, is presented in Table~\ref{table:metrics}.

\begin{table}[t]
    \centering
    \small
\begin{tabular}{c|c|c}
     \textbf{Metric Name} & \textbf{Type} & \textbf{Applied on}\\
     \hline
     Outlying\cite{wilkinson2005,wilkinson2006}       & Scagnostics  & LD  \\
     Skewed\cite{wilkinson2005,wilkinson2006}         & Scagnostics  & LD \\
     Clumpy\cite{wilkinson2005,wilkinson2006}         & Scagnostics  & LD  \\
     Sparse\cite{wilkinson2005,wilkinson2006}         & Scagnostics  & LD  \\
     Striated\cite{wilkinson2005,wilkinson2006}       & Scagnostics  & LD  \\
     Convex\cite{wilkinson2005,wilkinson2006}         & Scagnostics  & LD \\
     Skinny\cite{wilkinson2005,wilkinson2006}         & Scagnostics  & LD  \\
     Stringy\cite{wilkinson2005,wilkinson2006}        & Scagnostics  & LD \\
     Monotonic\cite{wilkinson2005,wilkinson2006}      & Scagnostics  & LD   \\
     ABW\cite{lewis2012ABW}            & Cluster separability & LD   \\
     CAL\cite{calinski1974CAL}            & Cluster separability & LD \\
     DSC\cite{sips2009DSC}            & Cluster separability & LD   \\
     HM\cite{gilad2004HM}             & Cluster separability & LD   \\
     NH\cite{paulovich2008NH}             & Cluster separability & LD  \\
     SC\cite{rousseeuw1987silhouettes}           & Cluster separability & LD  \\
     CC\cite{geng2005CC}            & Correlation btw distances & HD to LD  \\
     NMS\cite{kruskal1964NMStress}            & Stress           & HD to LD  \\
     CCA\cite{demartines1997CCA}            & Stress            & HD to LD   \\
     NLM\cite{sammon1969NLM}            & Stress            & HD to LD  \\
     LCMC\cite{chen2009}           & Small neighborhoods & HD to LD \\
     T\&C \cite{venna2006}          & Small neighborhoods & HD to LD  \\
     NeRV\cite{venna2010NeRV}           & Small neighborhoods & HD to LD   \\
     AUC$_{log}$RNX\cite{lee2015} & All neighborhoods & HD to LD  \\
\end{tabular}

\caption{List of measures used in our analysis. If the metric is said to be applied on LD, then it only measures the quality (or check patterns in) the visualization. These measures are said to capture how interpretable the visualization is. However, if it said to be applied from HD to LD, then it measures the accuracy of the DR process.}
\label{table:metrics}
\end{table}

Among the metrics in Table~\ref{table:metrics} that have not been presented in Section~\ref{subsec:RW_QM}, one can find the silhouette coefficient (SC), the correlation coefficient (CC), the non-metric stress (NMS), the curvilinear component analysis (CCA), the nonlinear mapping stress (NLM) and the neighbor retrieval visualizer (NeRV).

SC~\cite{rousseeuw1987silhouettes} is a classic metric in clustering that measures how clusters are separated from each other, versus how instances inside a same cluster are grouped together. This metric is similar to ABW, but diverges in its mathematical definition. CC~\cite{geng2005CC} is a metric that computes the correlation between the vector of all pairwise distances in the original dataset and the corresponding vector of pairwise distances in the visualization. NMS~\cite{kruskal1964NMStress}, CCA~\cite{demartines1997CCA} and NLM~\cite{sammon1969NLM} are three stress measures that are considered in our study. Stress measures have in common that they measure how well pairwise distances in the high-dimensional space are preserved in the low-dimensional space. Each of the three measures have their particularities. For instance, NMS~\cite{kruskal1964NMStress}, as a non-metric measure, does not compare pairwise distances directly, but their ranking.

Finally, NeRV~\cite{venna2010NeRV} is a metric based on information retrieval, in the sense that it translates the concepts of precision and recall to a measure similar to the Truthfulness and Continuity. Furthermore, similarly to the two sub-metrics of Truthfulness and Continuity, precision and recall are then combined by using, for instance, a simple mean. One particularity of NeRV is that it redefines the distances in the original dataset and in the visualization as probabilities, like $t$-SNE. It also contains a perplexity hyper-parameter that represents the size of the neighborhood to consider. In our experiments, NeRV perplexity has been fixed at 5. 

\begin{figure}[ht]
    \centering
    \includegraphics[width=0.9\linewidth]{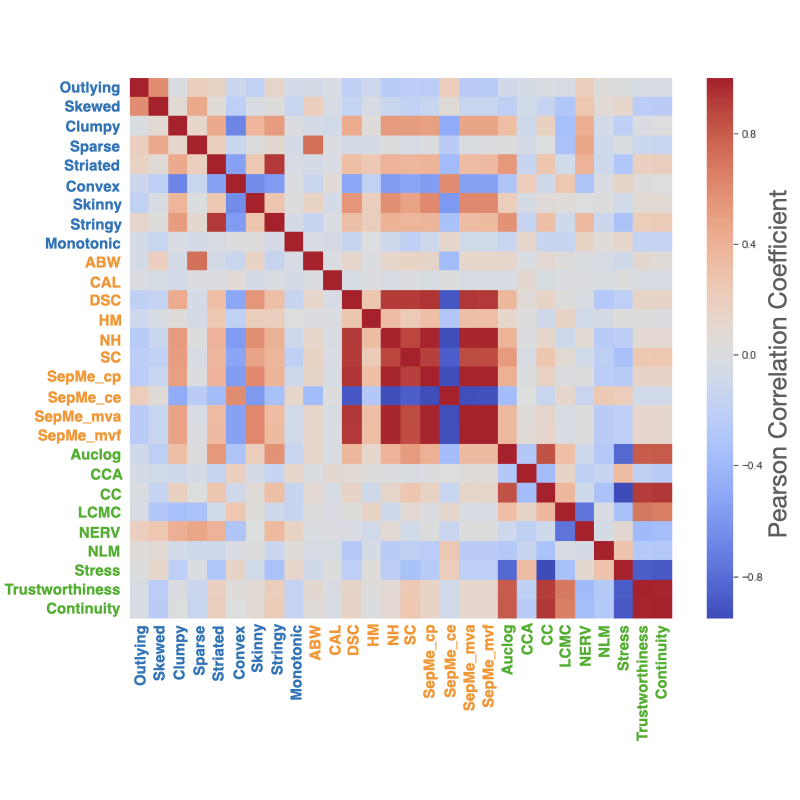}
    \caption{Correlation matrix of the 2 metric categories: interpretability (Scagnostics measures, in blue \& separability measures, in amber) and accuracy measures (in green).}
    \label{fig:metrics}
\end{figure}

In Figure~\ref{fig:metrics}, a correlation matrix heatmap of the calculated measures is presented. The separability metrics are all highly correlated. For this reason, for all analysis involved we decided to drop measures correlated at more than  95\%. Between pairs of highly correlated measures, the most popular one in each pair was kept. In consequence, the metrics dropped from further analysis were: $SepMe_{mvf}$, $SepMe_{mvt}$, Continuity, NH, and CC. Additionally we also removed ABW and CAL, as they were low variance features and carried low information.

\section{Modeling User Preferences}\label{sec:analysis}
We now present three ways to model our data, with incremental levels of detail:
\begin{itemize}
    \item The first model aims to loosely classify ``good'' and ``bad'' projections, as decided by users. In this case, the number of points awarded is not taken into account, only  whether the projection was crossed out or not. 
    \item The second model aims to linearly learn which projections are preferred by users, by answering the question ``Would projection A be preferred to projection B?''. 
    \item The third model seeks to provide a ranking of the projections. With this model, we specifically want to examine whether a nonlinear combination of the metrics can further improve the performance.
\end{itemize}

From all three models, we extract the most important metrics that help to predict user preferences. Although there are disagreements in the data, no data and no participants were discarded from the training process. Hence, the models were trained on noisy annotations where the same projection may have conflicting annotations. With this setup, we were able to take into account the subjectivity in the data. To conclude this section, we decide which of the three models should be used as part of our technique and proof-of-concept tool presented in Section~\ref{sec:case_study}. 

\subsection{Modeling Setup}

The evaluation of our models is operated on a leave-one-group-out basis. This is a cross-validation setup where the data is split into distinct groups and a model is trained on the collected preferences related to all groups but one. The remaining group is used as a test set. The process is repeated for all combinations of groups. Throughout our modelings, we mainly use the datasets in Table~\ref{table:datasets} as our groups. We call this procedure leave-one-dataset-out (LODO).

Given that different datasets are used to generate our projections, and that they have different degrees of complexity (see Table \ref{table:datasets}), it is expected that all our models vary slightly in performance from dataset to dataset. 
Furthermore, computing a prediction score for each group also enables us to  build a measure of prediction uncertainty on unseen data, by calculating the confidence interval over all test dataset results $\textbf{R}_{test}$.

\subsection{Model 1: Classifying Good and Bad Projections}\label{subsec:classification_XP} 

Model 1 is set up to learn the distinction between ``good'' and ``bad'' (i.e., misleading) projections. 
We classified a projection as good if a participant scored it with at least one heart, and as bad if it was crossed out by a participant. Figure~\ref{fig:trial} shows these two categories with scatterplots highlighted either in green (good) or in red (bad).

For this setup, the metrics were assigned as features and people's preferences were binarized to 1 (good) or 0 (bad). The data was fed to a random forest and evaluated on a LODO basis to determine the prediction performance for each dataset. We also cross-validated hyper-parameters to choose the best setup for each test fold of LODO. The setup was implemented using the Scikit-learn library in Python. The random forest with 200 decision trees of a maximum depth of 10 nodes was our best setup. In total, the model used 3664 annotated projections to learn. SHAPley values \cite{lundberg2020local} were used to explain the prediction for any instance $x_{i}$ as a sum of contributions from its individual feature values. This interpretation was similar to that of weights in a linear model, but in a model that can approximate more complex functions.

\begin{figure}
    \centering
    \includegraphics[width=\linewidth]{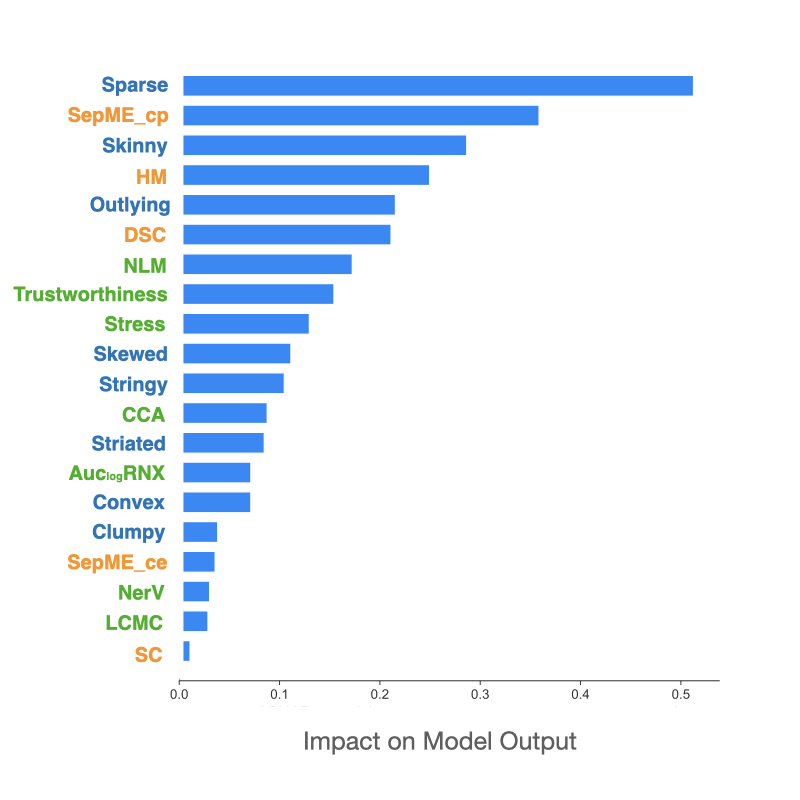}
    \caption{Top features used by Model 1. The features are listed in order of importance. The length of the bar represents the absolute impact on the model output. Sparsity is the most important feature, and its impact on the model output means that on average this feature can change the probability of being a ``good'' projection by 0.5.}
    \label{fig:binary_importance}
\end{figure}

The area under the receiver operator curve (AUC) metric was optimized in the LODO setup. This resulted in the predictive performance of 78.36\% with a confidence interval of $\pm$ 4.08\%. In terms of feature importance, Scagnostics~\cite{wilkinson2005,wilkinson2006} features such as Sparsity, Skinny and Outlying are the most important ones. Feature importance is summarized in Figure~\ref{fig:binary_importance}. For the majority of the tested projections, low Sparsity and high Skinniness increase the chances of a projection to be disliked by participants. This makes sense as projections selected by users as bad tend to be random projections, where points are scattered in the 2D visualizations, with no apparent meaning. 
An example of such a projection can be seen in the last position on the grid of Figure \ref{fig:trial}. A similar interpretation exists for very skinny projections such as the projections in the second and third to last places on the grid of Figure~\ref{fig:trial}.

\subsection{Model 2: Linear Preference Learning}\label{subsec:pref_learning_XP} 

For Model 2, we re-defined the problem as to linear preference learning. To do that, for each pair $(v_i, v_j)$ of visualizations in a dataset, the percentage of time $v_i$ is preferred over $v_j$ is considered. For instance, 90\% means that 90\% of the time, when $v_i$ and $v_j$ were presented in the same trial to users, $v_i$ received a larger number of hearts than $v_j$. Because the comparisons are aggregated to get percentages, the number of instances becomes 2268 for this dataset.

The goal of the linear preference learning model is then to linearly reconstruct the preferences between visualizations, based on the percentage of time a particular visualization has been preferred to another visualization. The advantage of linear models are their robustness to overfitting, as well as their interpretability. As for all experiments, the quality metrics are used as explanatory variables for predicting the preferences.

In order to do such predictions, Bradley-Terry models (BTm)~\cite{bradley1952BT} are used. BTm linearly combines features to derive probabilities of being preferred:
\begin{equation*}
    P(v_i > v_j) = \frac{e^{w_0 + w_1*m_{1,i} + ... + w_{23}*m_{23,i}}}{e^{w_0 + w_1*m_{1,i} + ... + w_{23}*m_{23,i}} + e^{w_0 + w_1*m_{1,j} + ... + w_{23}*m_{23,j}}},
\end{equation*}
where $w_0$, $w_1$, ..., $w_{23}$ are 24 weights to learn (one for each metric plus the intercept), and $m_{k, i}$ (resp. $m_{k, j}$) are the $k^{th}$ metric evaluated on the visualization $v_i$ (resp. $v_j$). We trained the BTm with a Lasso penalty in order to encourage sparsity among the weights. This enabled the model to obtain the lowest error that it could, while using the fewest quality metrics. Thus, the approach discarded metrics that have little to no effect in the prediction of participant preferences. For developing our BTm model, we modified the package BradleyTerry2 in R to include the Lasso penalty.

\begin{figure}
    \centering
    \includegraphics[width=\linewidth]{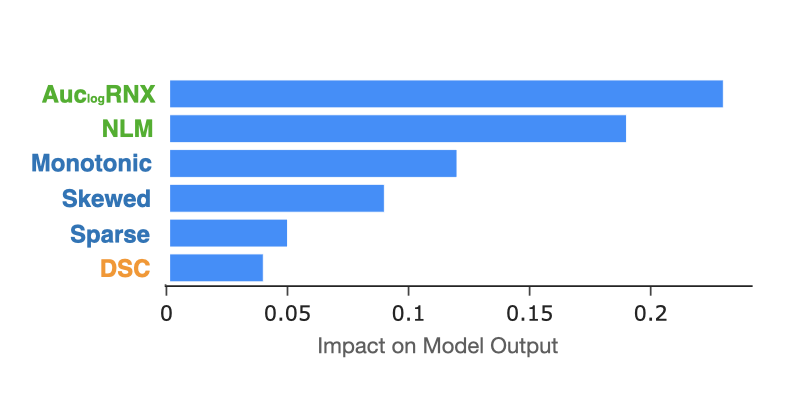}
    \caption{Top features used by Model 2. Given the Lasso regularization, this model uses only 5 features compared to the other 2 models.}
    \label{fig:exp2_importance}
\end{figure}

The absolute value of the metric weights that have been found after learning a sparse BTm on our preference data are presented in Figure~\ref{fig:exp2_importance}. The accuracy of the BTm is 62.3\%, with the 95\% confidence interval being [58.39\%, 66.22\%]. The accuracy is obtained by counting the number of time the model is right when it says $v_i > v_j$, over the total number of predictions. To obtain accuracy on data that have not been used for training, the LODO strategy has been used. The final accuracy is the mean of the test accuracy scores of the 11 involved datasets. This way, the reported final accuracy offers some guarantees on the use of the presented sparse linear model on new datasets. If only the data where users strongly agree on good and bad visualizations (at least 80\% of agreement) is used, the accuracy becomes 65.93\% [61.42\%, 70.43\%]. The $\lambda$ balancing the importance given to the error and the Lasso penalty was $0.021$ for a BTm learned on the whole dataset.

\begin{figure}[ht]
    \centering
    \includegraphics[width=\linewidth]{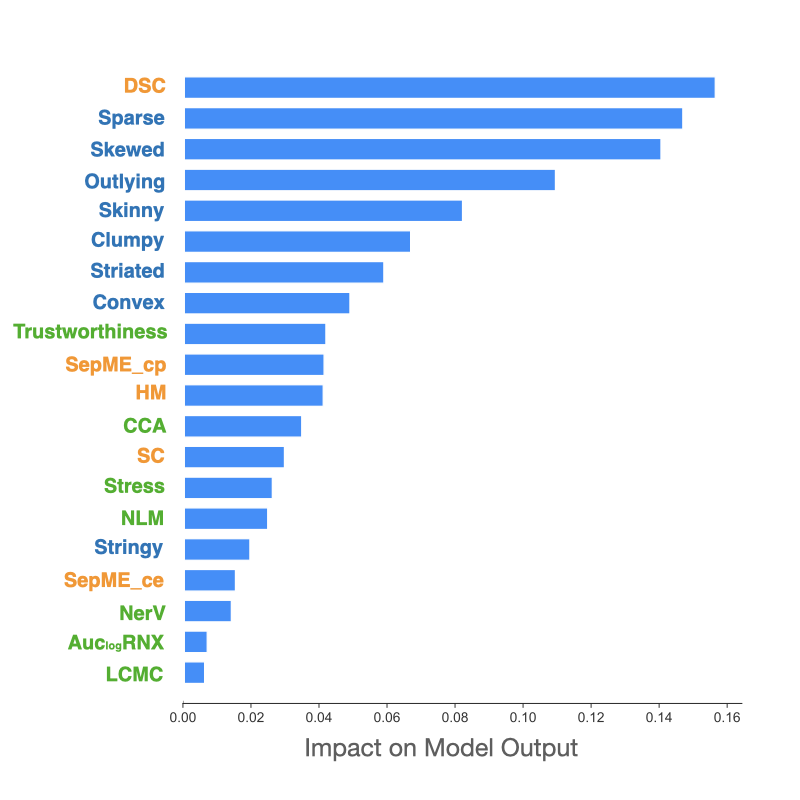}
    \caption{Ranked features of Model 3. Three out of the top five features (DSC, Sparse and Skinniness) are in alignment with features from Model 2.}
    \label{fig:ranking_importance}
\end{figure}

\subsection{Model 3: Nonlinear Ranking of Projections}\label{subsec:ranking}

In our final setup (Model 3), like for Model 2, we output a measure of how good each projection is.
This makes it possible to answer the question ``By how much is projection A better than projection B?''. This measure acts as a popularity score and can also be used to compare if the projections generated for some datasets have a higher quality than for other datasets. 

As opposed to Model 2, we chose for Model 3 a nonlinear model to exploit more complex relationships among the metrics and potentially increase our performance. We implemented a boosted tree ensemble to both rank the projections of each dataset. 
Boosted tree ensembles are learning methods that can be used for classification, regression, and learning-to-rank tasks~\cite{burges2010ranknet}.

Given that boosted trees are state-of-the-art models in supervised learning for tabular data, we expect that exploiting the nonlinear relationships between our features could lead to performance improvement. That is, we would like to know whether a nonlinear combination of our features, unlike the one mentioned in Equation~\ref{alpha_beta}, can lead to better results. 

\begin{figure*}[ht]
    \centering
    \includegraphics[width=\linewidth]{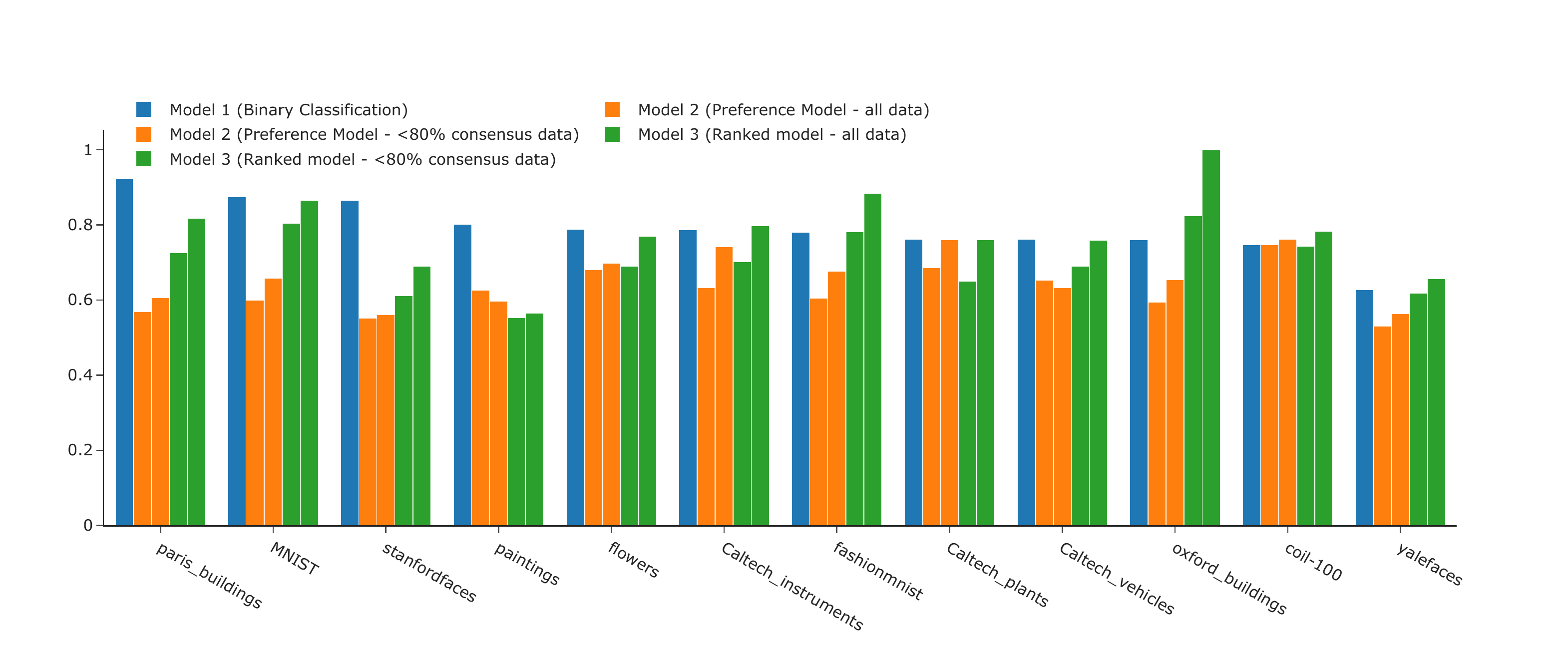}
    \caption{Overview of the performance of the 3 experiments evaluated for each dataset separately. The first experiment (binary classification) is the best performing as it is concerned by an easier task. }
    \label{fig:overview}
\end{figure*}
Model 3 is fed with projection lists that are sorted according to the hearts awarded by participants. 
As such, the model learns to rank the projections from the 3664 instances, but sorted into 458 groups of 8 projections, as they were initially ranked by our participants. Model 3's objective is then to create a ranking for a new, unseen, dataset of projections. This set of projections can be of any length, not just 8 projections, and the model learns to minimize the number of incorrect pairwise comparisons, as described by the LambdaMART algorithm~\cite{burges2010ranknet}. Following cross-validation of our hyper-parameters, our model was trained using 15 sequentially trained decision trees. The learning rate used in the setup was 0.3 and the maximum depth of each decision tree involved was 5. The setup was implemented using the XGboost library in Python.  

The LODO error is calculated the same way as in Model 2. Overall, the accuracy is 70\%, with a confidence interval (CI) of $\pm 4.4\%$. When the LODO error is calculated only for comparisons where there was a strong agreement, such as 80\% agreement, the accuracy increases to 78.09\%, with a CI of $\pm 6.5\%$.

\subsection{Discussion}

\textbf{Performance on Unseen Datasets:} Figure~\ref{fig:overview} displays the breakdown of performance accuracy for each dataset. Unsurprisingly, the model performs better on datasets which were rated as easier and with more consensus (see Table \ref{table:datasets}). Given the LODO errors from all experiments, we can establish with a CI of about 95\% that our models are able to generalize to new image datasets. 

\textbf{Performance of Features/Metrics:} Figures~\ref{fig:binary_importance}, \ref{fig:exp2_importance}, \ref{fig:ranking_importance} present the feature importance in our three models. In other words, they show which of the existing metrics (= features) are important for modeling human preferences of DR projections.
All three experiments show similar trends in terms of feature importance. Scagnostics features~\cite{wilkinson2005,wilkinson2006}, like Sparsity, Skewed and Skinny, alongside separability metrics, like DSC, are in all cases among the top 5 most important features. Features such as Sparse, Skewed, and Outlying are used to detect bad projections. These features tend to be high for projections where the positioning of the points appears random or uniformly distributed. These were universally disliked by humans, which can be seen in Figure~\ref{fig:dr_descriptive}, where the Gaussian random projection (GRP) was the most disliked DR technique. Previous work from Lehman et al.~\cite{lehmann2015study} also identified a subset of Scagnostics measures, namely stringy and striated, as measures which can be used to ``early reject'' projections that are not understandable for users. 

\begin{figure}[ht]
    \centering
    \includegraphics[width=0.9\linewidth]{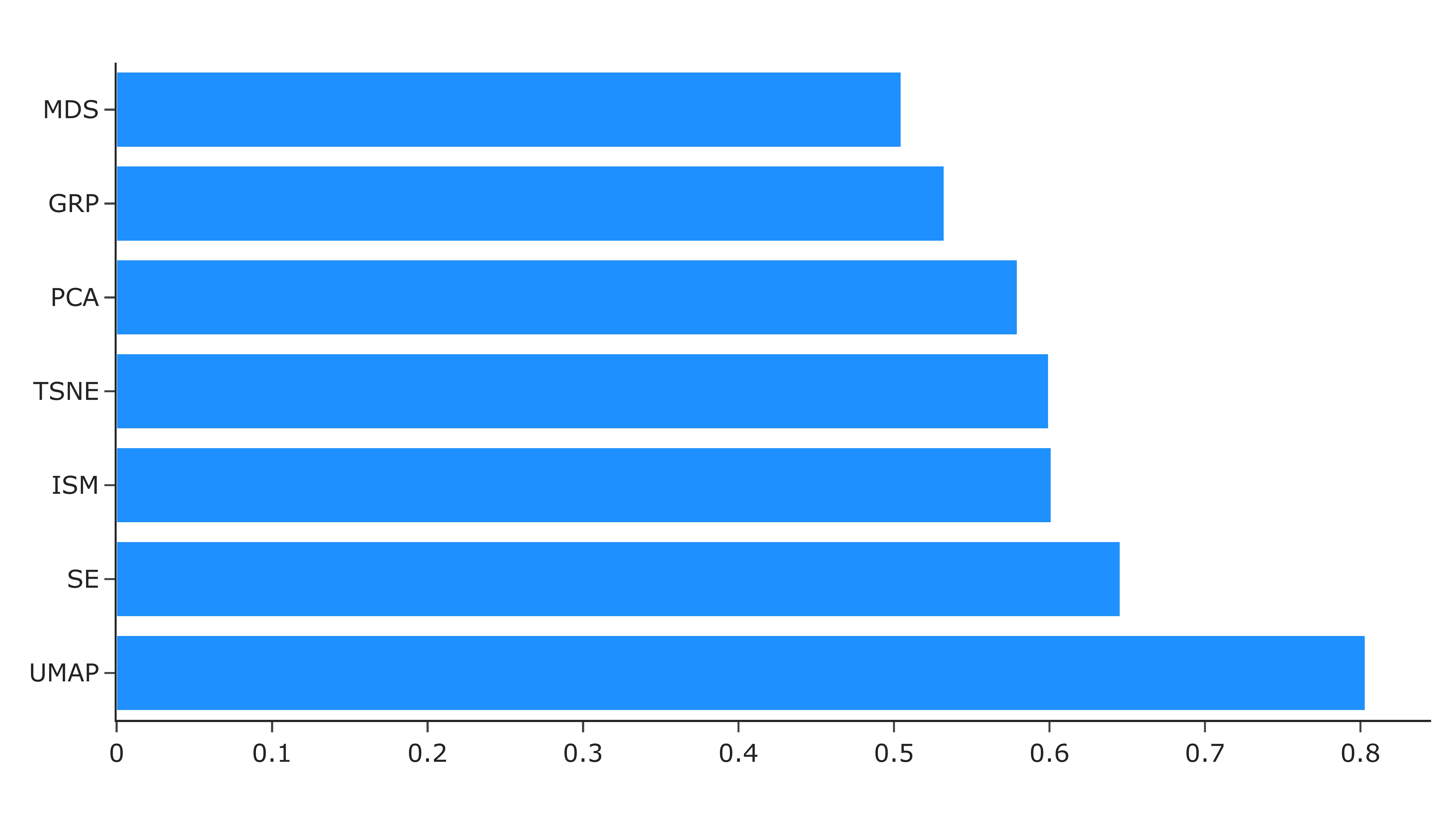}
    \caption{Overview of the performance of the first experiment evaluated for each DR technique separately.}
    \label{fig:lodro}
\end{figure}
In Model 2, the accuracy metrics NLM and AUC$_{log}$RNX have a large impact on the model (see Figure~\ref{fig:exp2_importance}). They are not compensating each other, as removing from the available metrics one of the two leads to a new model with a reduced performance. The higher importance of AUC$_{log}$RNX and the reliability of DSC, among cluster separability measures, to assess user preferences are aligned with similar experiments in the literature~\cite{bibal2016NIPS}. Indeed, all models use separability features, such as DSC, to detect the presence of semantically relevant clusters. A high DSC measure is a strong indicator of a liked projection. This is in line with the quantitative evaluation undertaken by Sedlmair et al.
~\cite{sedlmair2012taxonomy}, which highlights class separability as one of the most important tasks people perform on DR. While Scagnostics measures are used like the other models, we can also see the Scagnostic measure Clumpy, which identifies clusters regardless of their semantic composition. This points to the fact that people prefer looking more specifically for clusters that make sense semantically. 

\begin{figure*}[!t]
    \centering
    \includegraphics[width=0.9\textwidth]{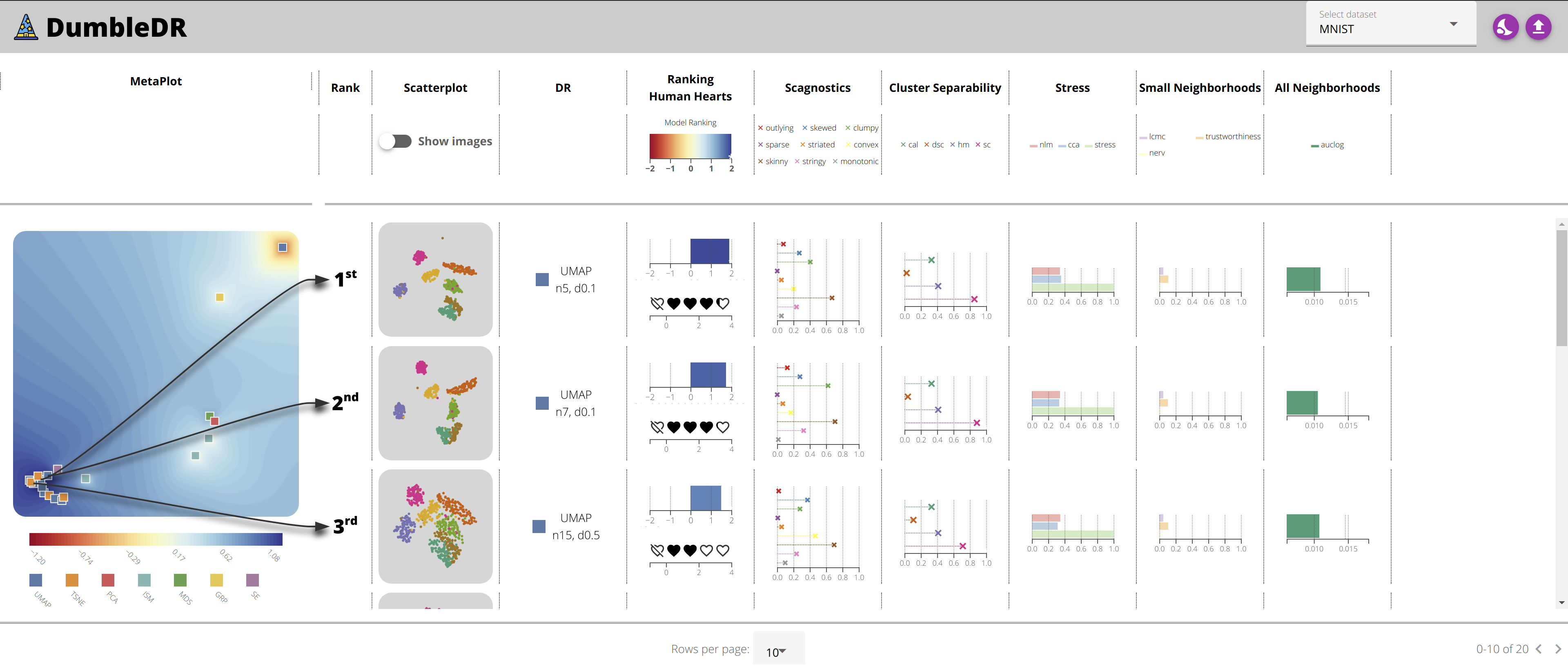}
    \caption{Screenshot of the tool for ranking projections. The projections in the scatterplots column are ranked using Model~3. On the left of the ranking, a metamap shows similar  projections (w.r.t. the quality metrics) close together and dissimilar ones far apart. The blue (resp. red) zone represent good (resp. bad) projections w.r.t. Model 3 scores. On the right of the ranking, the average number of hearts given by participants is shown, as well as quality metric values.}
    \label{fig:tool}
\end{figure*}

All three models show that metrics from both, the VIS and the ML communities are important.
In addition to Scagnostics and cluster separability measures for detecting bad projections, our models also rely on accuracy measures to find accurate projections among the ones that contain readable patterns. 
This result logically stems from the fact that users do not pay much attention to the semantics inside visualizations if the instances do not form readable patterns. At the same time, users will not select visualizations containing clear patterns, which make no sense according to the high-dimensional data though (e.g. clear clusters consisting of random images).

\textbf{Performance of DR techniques:} A potential bias spanning from the type of datasets selected (i.e. image collections) is that linear techniques such as PCA get rated down. Given the fact that images lie on a nonlinear manifold in the high-dimensional space, it makes sense that linear DR methods such as PCA underperform in comparison to UMAP or $t$-SNE.

To evaluate the generalization to new DR techniques, a leave-one-dimensionality reduction-out (LODRO) error is calculated for Model 1. Rather than splitting by dataset during our cross-validation, as in LODO, we train to detect ``good'' and ``bad'' projections by considering all dimensionality reduction techniques but one. The LODRO procedure allows us to check if our analysis applies to new, unseen, DR techniques. 

Overall, our generalization error to new DR techniques is settling at 59.8\%, with a confidence interval of $\pm$ 9\%. Figure~\ref{fig:lodro} breaks down our results per DR technique for this analysis. GRP and MDS have the worst generalization error. The explanation can be that these particular methods bring very different projections than the other DR techniques. However, users in our study graded the projections resulting from GRP, SE and some UMAP configurations as universally bad across all datasets (see Figure~\ref{fig:dr_descriptive}). Users have even commented about how these projections appear to be random. However, visualizations that appear to be random to the human eye have in fact a very different quality according to quality metrics, meaning that bad projections are not all bad in the same way. On the flip side, most configurations of UMAP, which is one of the newest proposed DR techniques in the literature, generalize very well. An interesting future direction is to assess which minimal set of dimensionality reduction techniques could be jointly used to train models such as ours in order to ensure that the resulting projections are diverse enough to generalize well. 

It should be noted that the LODRO strategy cannot be easily applied for the Models 2 and 3, since, in these setups, we would require more DR techniques, and more than 20 total projections per dataset in order to achieve significant results.

\subsection{Model Selection}
We now would like to select one of these models for our tool DumbleDR.
Theoretically, all three models can be used to reliably make predictions for the introduced tasks. The accuracy of the two nonlinear models 1 and 3 is, with values above 75\%, slightly higher than the accuracy of Model 2. If a user only wishes to filter out bad projections, we recommend Model 1, as it has the higher accuracy on our datasets. However, given that we have set out to rank predictions, we have selected Model 3 to use in our tool DumbleDR. The selection was made based on accuracy performance criterion as with Model 3 better results shall be expected in general.

\section{DumbleDR}\label{sec:case_study}

This section presents a web-based visual analytics tool, named DumbleDR\footnote{The tool is available at \url{https://renecutura.eu/dumbledr/}}, containing an implementation of Model 3, in order to showcase how to use our technique. While Model 3 is used in the tool, the other two models could be similarly plugged into DumbleDR. The following sections present the tool in more details, as well as two case studies showing the analysis of two new datasets with our proposed model. 

\begin{figure*}[!t]
    \centering
    \includegraphics[width=\textwidth]{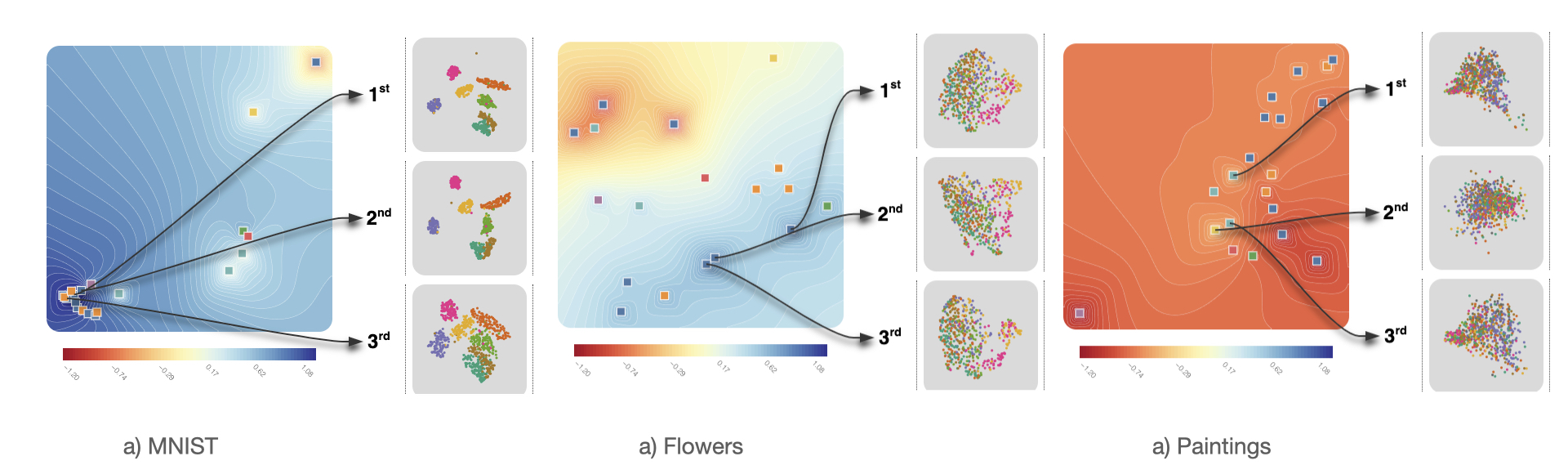}
    \caption{Top 3 best projections, as scored by Model 3, for three of the datasets we have collected: MNIST handwritten digits, photos of flowers and, Art UK paintings. For each dataset, we provide metamaps where each square represents a projection for the particular dataset. The metamaps are calculated by applying dimensionality reduction on the quality metrics space and the color-coded contours represent the ranking score predicting human preferences. Well-liked projections tend to be generated from the same neighbourhood in the metamap manifold. The spread of the ranking score varies across the three datasets, informing the user that the best projections for a dataset are not necessarily great quality. For instance, the MNIST dataset produces stronger candidates than the paintings dataset. }
    \label{fig:metamaps}
\end{figure*}
\subsection{Presentation of the Tool}

Our tool aim is to demonstrate how users can make sense of our model outputs on novel datasets. DumbleDR (i) takes as input new datasets, (ii) computes a range of projections and their associated metrics, and (iii) outputs a ranking of the projections, along with numerous statistics about their dataset quality. The tool uses JavaScript, specifically the Druid package~\cite{yang2014druid}, to compute all projections and metrics, and D3~\cite{bostock2011d3} for visualizations.

Figure~\ref{fig:tool} shows a screenshot of DumbleDR. The tool can either be used to explore new datasets or to check our experimental results. After selecting a precomputed dataset or uploading a new one (on the top-right corner of the screen), projections and their respective quality metrics are computed. Without additional training required, Model 3 will output a score for each projection of the dataset. The output score is a real number which can be positive or negative. The higher the number, the better the projection is. The resulting projections are ranked in accordance to this output.

When uploading a novel dataset to our tool, the tool first computes a number of projections, then the associated quality metrics, and finally, the ranking. Of these three tasks, computing the metrics, in particular the accuracy ones, is the most expensive operation. This is because accuracy metrics use the high-dimensional space to compute distance-based neighbors in order to compare them with low-dimensional neighbors. If Scagnostics metrics take less than a minute to compute for 40 projections of a dataset, separability measures take minutes, and accuracy measures can span hours.

On the left of the screen, in Figure~\ref{fig:tool}, a metamap shows the similarity between the projections created based on a selected dataset. This metamap is a UMAP projection over the metrics calculated for each DR projection from the original dataset. This approach was first introduced by Cutura et al. in their system VisCoDeR~\cite{cutura2018viscoder}. The colors in the metamap represent the ranking score of the visualizations: from dark blue for great visualizations, according to Model 3, to dark red for low-ranked ones.

The spread of the ranking score outputed by Model 3 varies from dataset to dataset. This can be seen in Figure~\ref{fig:metamaps}, which shows the metamap of three datasets, MNIST, Flower photography, and ART UK paintings, and the corresponding top three projections. The information encoded in the metamap contours can be used to deduct that the projections from MNIST are rated high across the ranking (large blue zone) and, therefore, that lower ranked projections can also be considered. For the paintings dataset, however, only few projections are good (large red zone), and Model 3 helps to find these good projections. The flower dataset, in the middle of Figure~\ref{fig:metamaps}, is balanced, as it contains both good and bad projections. In conclusion, not all produced projections are equal in terms of quality, and our ranking score, a combination of the metrics based on user preferences, is indicative of that. 

On the right of the metamap in the tool (see Figure~\ref{fig:tool}), the ranking of visualizations is presented, with arrows linking them to their position on the metamap. The DR column, which is on the right of the scatterplots column, provides all information about the embeddings used to obtain the visualization, along with their parametrization when relevant. The other columns show other information like the average number of points the visualization obtained during the user experiment, if available, and the scores from the individual quality metrics. The user can compare the ranking score with the average points awarded by people for each projection during the user study. 

\subsection{Use Cases}

In this section, we present two use cases on two distinct and novel datasets that were not used in the user study or the previous analysis. The objective of the use cases is to present how to use our tool, and therefore the implementation of Model~3, to obtain projections ranked by quality.

\subsubsection{Use Case 1: The Pets Dataset}

In a first use case, let us consider a user who wants to get a visualization of the pets dataset~\cite{parkhi2012pets}. This dataset contains 38 classes of various breeds of cats and dogs. All previous datasets used in this analysis contained a maximum of 7 classes. The reason was to avoid overwhelming users during our study. In this case study, we aim to see if our technique can be successfully applied on datasets with a much higher number of classes.

\begin{figure}[!ht]
    \centering
    \includegraphics[width=0.45\textwidth]{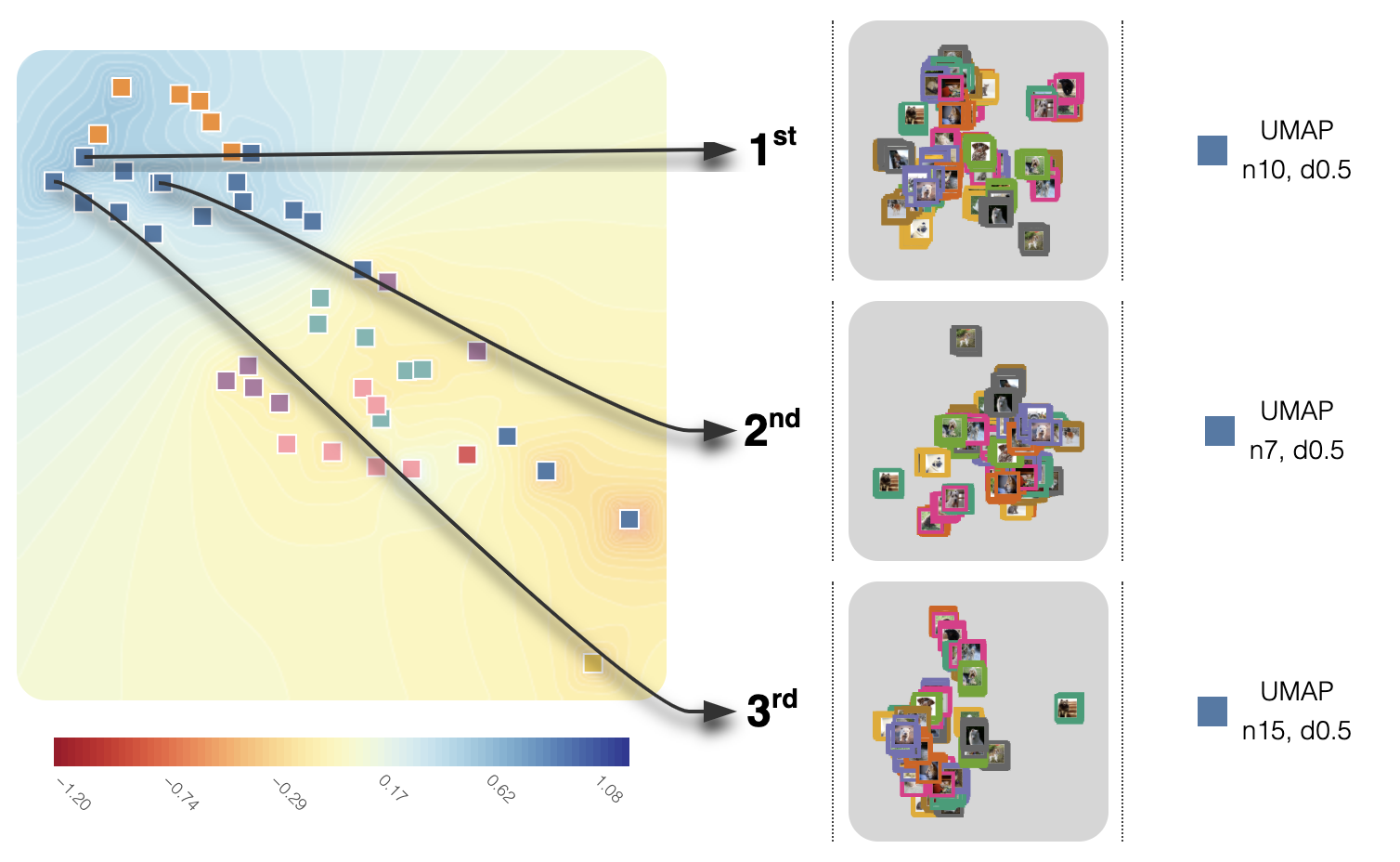}
    \caption{Top 3 projections given by our tool on the pets dataset. The ranking is provided by Model 3 and shows that UMAP with some particular parametrizations offers visualizations of good quality.}
    \label{fig:pets_UI}
\end{figure}

Figure~\ref{fig:pets_UI} shows the best projections that can be obtained on this dataset. DumbleDR with Model 3 shows that UMAP can provide good visualizations of the pets dataset, and it also provides the parametrization to obtain these good UMAP projections. Getting the right parametrization is essential, as the worst visualizations, in the most red parts of the metamap, are also UMAP projections, albeit with different parametrizations.

\subsubsection{Use Case 2: Selecting Good Metamaps}

As a second use case we discuss the right choice of metamaps for comparing projections. Defined by Cutura et al.~\cite{cutura2018viscoder} and used in this paper, metamaps are projections of projections. They are used to compare projections, and find similar or dissimilar projections, encoded by the distance in between points (projections) in the metamap. 
Another example use of a metamap would be to collect the most different projections in order to get different views of the same data. In order to do that, one would compute hundreds of projections, produce the metamap, and consider the  projections that are the most distant from another. However, in all applications, if the metamap used is not accurate or not readable, no insight can be extracted from it.

We produced the metamaps for this use case by taking all the projections generated by our datasets (Table~\ref{table:datasets}), computing the metrics for each of them, and finally applying dimensionality reduction techniques on them. Therefore, each point in the visualization is a metamap of projections. The separability metrics took as labels the dataset associated with the projection. We followed the same procedures as described in Section~\ref{sec:data_collection}. 

\begin{figure}[!ht]
    \centering
    \includegraphics[width=0.45\textwidth]{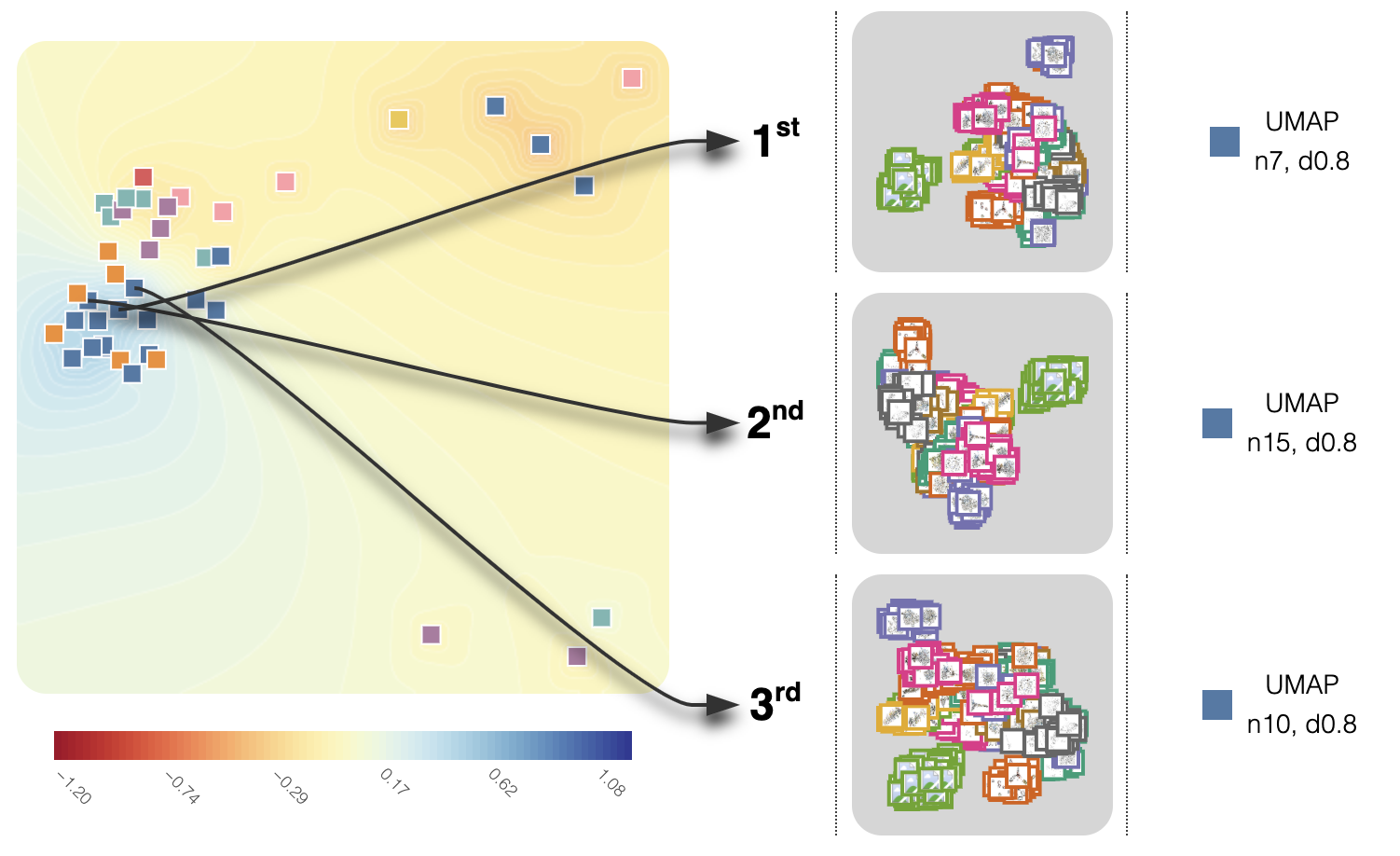}
    \caption{Top 3 projections given by our tool on the set of metamaps. The ranking is provided by Model 3 and shows that UMAP with some particular parametrizations offers visualizations of good quality.}
    \label{fig:metamaps_UI}
\end{figure}

Figure~\ref{fig:metamaps_UI} shows the best metamaps according to the combination of metrics from Model 3. As for the previous example, UMAP provides the best metamaps when a certain parametrization is chosen. The metamaps shown by the tool for the other datasets are indeed produced by UMAP with the neighbours set to 7, and minimum distance of 0.8, which was selected by our algorithm as the best. The best metamaps showed that the projections are loosely clustered in accordance to dataset, rather than DR technique.

By using our tool and the combination of quality metrics implemented in it (i.e. Model 3), users can upload their dataset and get the techniques and parametrizations that provide the best projections. This eases the cumbersome process of (i)~running many different DR techniques, (ii)~testing many different parametrizations, (iii)~finding, implementing and understanding many different quality metrics, and most importantly, (iv)~selecting the best projection according to these quality metrics. Indeed, regarding (iv), our tool provides the combination of quality metrics that best predicts what users would consider as being a projection of quality.

\section{Limitations \& Future Work}\label{sec:discussion}

As all research, our work comes with a set of limitations, which specifically attain to the modeling approach (Section~\ref{sec:analysis}) based on inherently imperfect human subject data (Section~\ref{sec:data_collection}). 

\subsection{On the Existence of Misleading Projections}
An obvious concern one can have is that human subjects can select visually appealing projections that are nonetheless wrong with respect to the high-dimensional data (false positives). Based on the breadth and expertise of our user sample, as well as on the intrinsic availability to information regarding the high-dimensional space (thumbnail images), we are confident that if any such ``false positives'' existed, they would have been caught and marked as misleading or bad. Given that, our different models show that the majority of projections flagged as bad by participants can be detected using Scagnostics and separability measures. Given that no accuracy metric is needed for spotting these bad projections, it rises the question of whether projections where meaningful clusters are formed in the visualization, even though these clusters do not exist in the high-dimensional space, are even possible.

\subsection{On the Limited Breadth of Dataset Types} 

A weakness introduced by our study is that we only use image-based datasets. We did so, as images give a natural anchor into the high-dimensional space, which was essential for our purpose (see above). For this reason, we can only speculate that 1) users maintain their preferences for different dataset types, and, 2) that the metrics applied on different dataset types generate a similarly distributed metric dataset. An interesting future research direction would consist of extending the study with additional datasets of different types, such as tabular or text data as used in the quantitative survey from Espadoto et al.~\cite{espadoto2019towards}. Such studies will necessitate adequate low-dimensional representations of the high-dimensional tabular or text space, similar to the image thumbnails for image datasets. Glyphs might be an interesting route here.

\subsection{On the Number of DR Techniques and Quality Metrics} 

To the best of our knowledge, the DR techniques and quality metrics presented in this paper are a representative set of what is popular in the literature. However, one can argue that DR techniques and quality metrics that are not yet popular are not used. Even more, one can argue that new DR techniques and quality metrics can be invented in the future. While this is true, one contribution of this paper is also to present a framework on the use of quality metrics to predict user preferences in projections. This means that new metrics can be plugged into our framework so that a new combination is automatically learned and then analyzed without needing additional user feedback. Similarly, the combination can be re-trained on projections produced by new DR techniques, which would require a new user evaluation of these projections.

\subsection{Predicting User Behavior when Comparing Projections}

Potential future work can consist of using the characteristics from users in our models to derive a different combination of metrics per user profile. This could be done, for instance, by using variants of BTm. Indeed, the BTm presented in this paper can be used to analyze how user characteristics influenced their comparisons of projections. While BTm was used to predict the preferences based on features of the compared objects (the projections), BTm can also be used to predict the preferences based on the features of the ones that stated their preferences.

\section{Conclusion}\label{sec:ccl}

This paper tackles the problem of assessing the quality of dimensionality reduction (DR) visualizations using metrics from two research communities. The first group of metrics comes from the machine learning (ML) community and is used to assess the faithfulness of visualizations w.r.t. the high-dimensional (HD) data. The second group of metrics comes from the information visualization (VIS) community and is used to quantify the presence of readable patterns in the visualization. We proposed combining these different metrics in order to identify the important ones and draw conclusions for the two communities. We implemented a series of machine learning models to predict human preferences and examine to what extent metrics from both communities are used. The final model (Model 3) achieves  78.09\% accuracy in predicting both well-liked and misleading projections. Furthermore, Model 3 was implemented in a tool to demonstrate the capabilities of the proposed technique to highlight high quality projections.

In all three models, Scagnostics and separability measures from the VIS community have a large impact for predicting user choices. In particular, these metrics were able to easily discriminate between visualizations deemed good or bad by users. It seems that accuracy metrics from the ML community are secondary, but they make it possible to discriminate between accurate and misleading visualizations with readable patterns.

\section*{Acknowledgements}
Funded by the Deutsche Forschungsgemeinschaft (DFG, German Research Foundation) –- Project-ID 251654672 –- TRR 161 (Project A08).

\bibliographystyle{abbrv-doi}

\bibliography{biblio}

\begin{IEEEbiography}[{\includegraphics[width=1in,height=1.25in,clip,keepaspectratio]{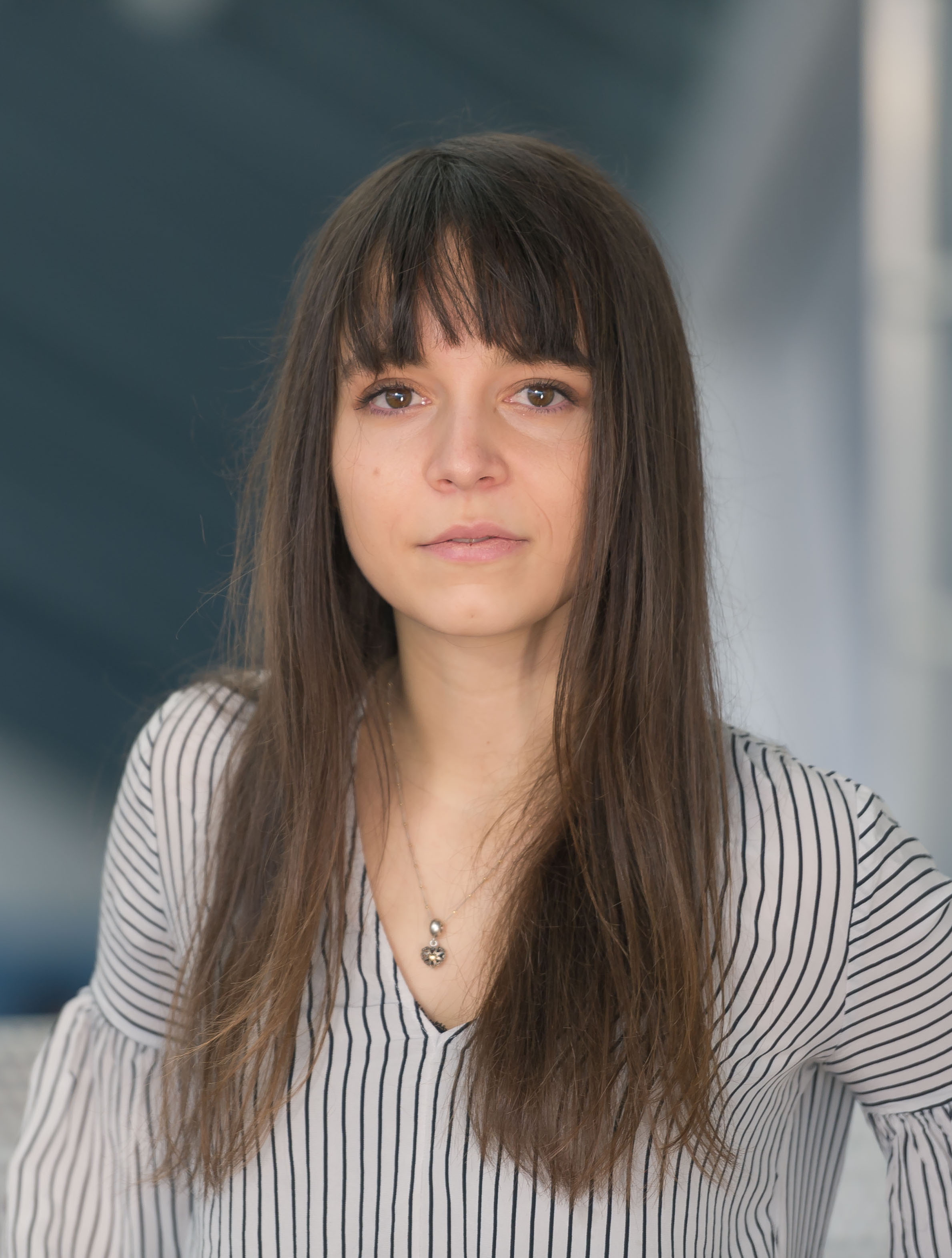}}]{Cristina~Morariu} was a researcher at the University of Stuttgart (Germany) in the group of  Professor Michael Sedlmair until October 2020. Currently, she works as a Machine Learning Scientist at Amazon. She received an M.Sc. degree in Operational Research with Data Science from Univeristy of Edinburgh in 2017. 
\end{IEEEbiography}

\begin{IEEEbiography}[{\includegraphics[width=1in,height=1.25in,clip,keepaspectratio]{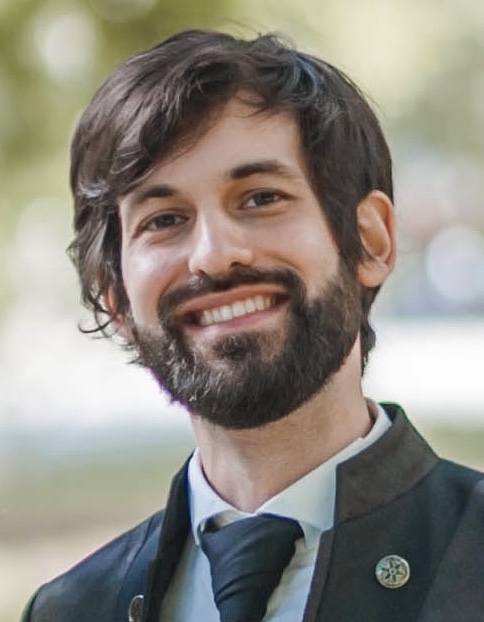}}]{Adrien~Bibal} is a postdoctoral researcher at the University of Namur (Belgium). He received an M.S. degree in Computer Science and an M.A. degree in Philosophy from the Universit\'{e} catholique de Louvain (Belgium) in 2013 and 2015 respectively. His Ph.D. thesis in machine learning, completed in 2020 at the University of Namur (Belgium), was on the interpretability and explainability of dimensionality reduction mappings.
\end{IEEEbiography}

\begin{IEEEbiography}[{\includegraphics[width=1in,height=1.25in,clip,keepaspectratio]{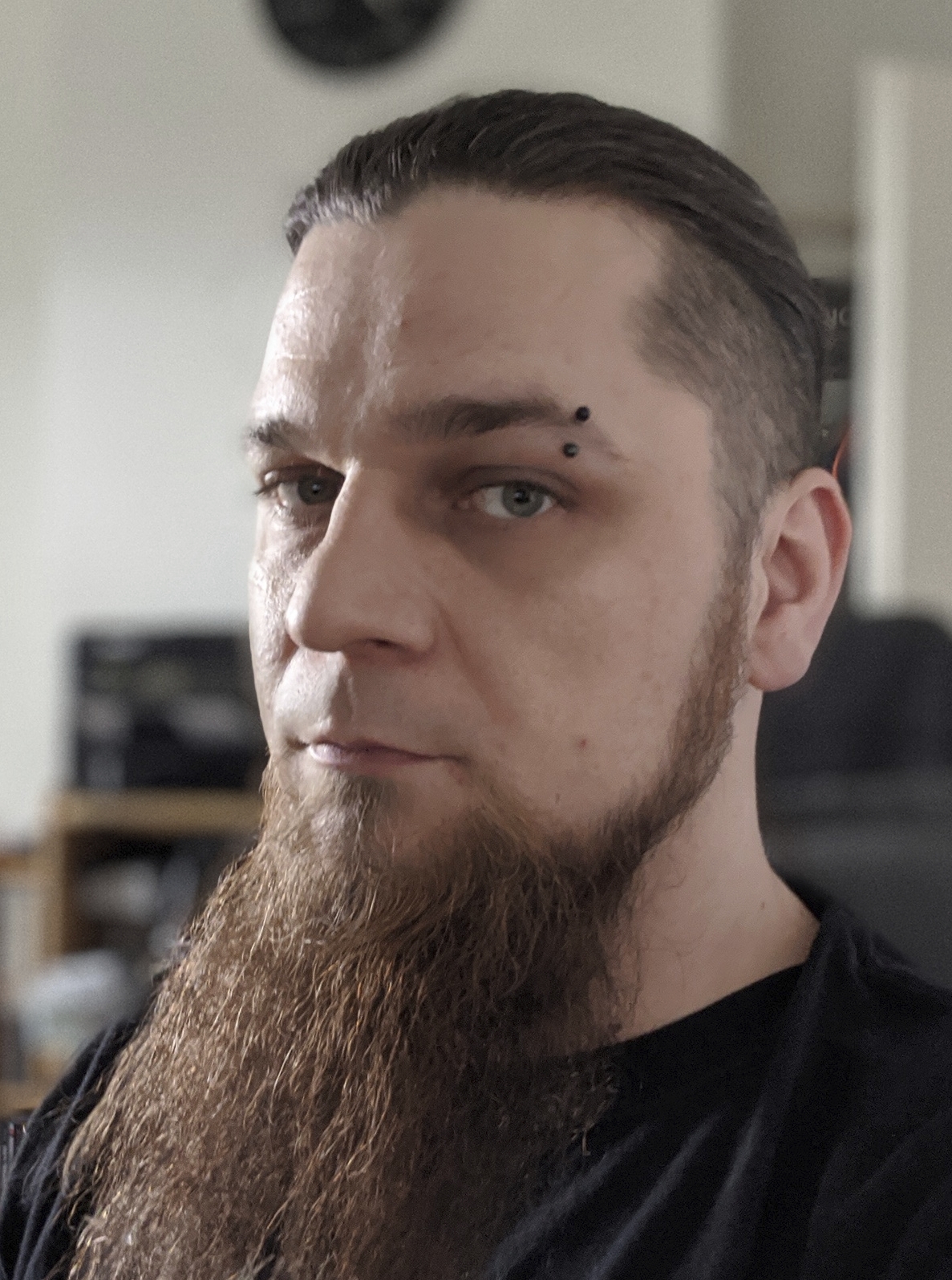}}]{Rene~Cutura} is a researcher at TU Wien (Austria) and is part of the research group of Professor Michael Sedlmair at the University of Stuttgart (Germany). He received a degree at the University of Vienna (Austria) in teaching mathematics and informatics. His research interests focus on visualizations of high-dimensional data, and dimensionality reduction.
\end{IEEEbiography}

\begin{IEEEbiography}[{\includegraphics[width=1in,height=1.25in,clip,keepaspectratio]{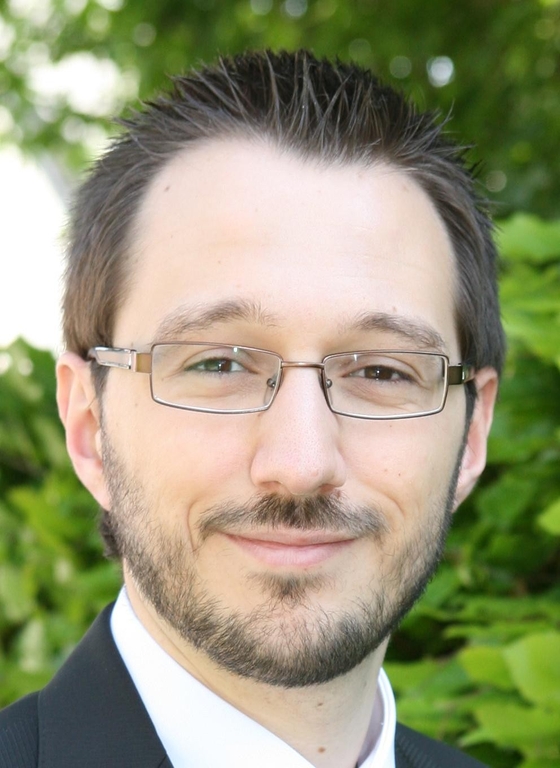}}]{Beno\^{i}t~Fr\'{e}nay} is associate professor at the Universit\'{e} de Namur.  He received his Ph.D. degree from the Universit\'{e} catholique de Louvain (Belgium) in 2013. His main research interests include interpretability, interactive machine learning, dimensionality reduction, label noise, robust inference and feature selection. In 2014, he received the Scientific Prize IBM Belgium for Informatics for his PhD thesis on Uncertainty and Label Noise in Machine Learning.
\end{IEEEbiography}

\begin{IEEEbiography}[{\includegraphics[width=1in,height=1.25in,clip,keepaspectratio]{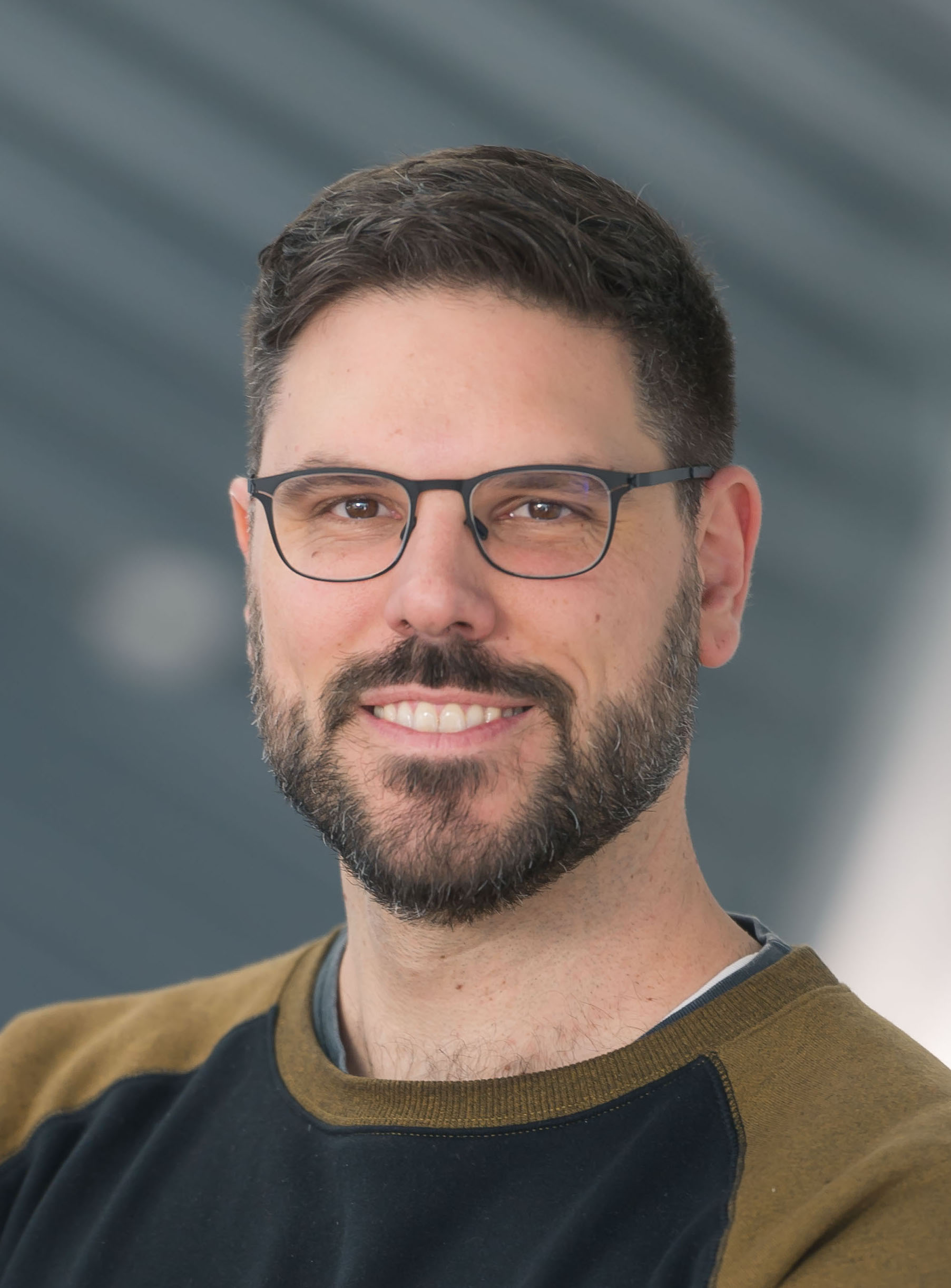}}]{Michael Sedlmair} is a professor at the University of Stuttgart
and is leading the research group for Visualization and Virtual/Augmented 
Reality there. He received his Ph.D. degree in Computer Science from the
Ludwig Maximilians University of Munich (Germany), in 2010. His  research interests focus on
visual and interactive machine learning,
perceptual modeling for visualization,
 immersive analytics and situated visualization,
  novel interaction technologies, as well as the 
methodological frameworks underlying them.
\end{IEEEbiography}


\end{document}